\documentclass[runningheads]{llncs}
\RequirePackage{silence}
\usepackage{amssymb, graphicx, amsmath}
\usepackage{comment}
\usepackage{color}
\usepackage{url}
\usepackage{hyperref}
\usepackage{overpic}
\usepackage{multirow}
\usepackage{textcomp}
\usepackage{gensymb}
\usepackage{scalerel}
\usepackage{overpic}
\usepackage{float}
\usepackage{multirow}
\definecolor{tabfirst}{rgb}{0.7, 1, 0.7} %
\definecolor{tabsecond}{rgb}{0.85, 1, 1.0} %
\definecolor{tabthird}{rgb}{1, 1, 0.7} %
\usepackage{colortbl}
 \usepackage{cleveref}
  \usepackage{booktabs}
\usepackage{cuted}

\makeatletter
\renewcommand\@fnsymbol[1]{\ensuremath{\dagger}}
\makeatother

\newif\ifreview
\reviewfalse
\ifreview
	\usepackage{lineno}

	\linenumbers
\fi

\begin{document}

\def\SubNumber{31}

\def\GCPRTrack{Main Track}

\title{VisDom: Sparse Novel View Synthesis with Visible Domain Constraint}

\ifreview
	\titlerunning{GCPR 2026 Submission \SubNumber{}. CONFIDENTIAL REVIEW COPY.}
	\authorrunning{GCPR 2026 Submission \SubNumber{}. CONFIDENTIAL REVIEW COPY.}
	\author{GCPR 2026 - \GCPRTrack{}}
	\institute{Paper ID \SubNumber}

\else

	\author
    {Mariia Gladkova*\inst{1,2}\thanks{Corresponding Author; * equal contribution}
    \and
	Tarun Yenamandra*\inst{1,2}
    \and
	Edmond Boyer
    \and
    Robert Maier
    \and
    Tony Tung
    \and
    Daniel Cremers\inst{1,2}
    }

	\authorrunning{M. Gladkova et al.}

	\institute{\textsuperscript{1}TU Munich, \textsuperscript{2}MCML}
\fi

\maketitle              %
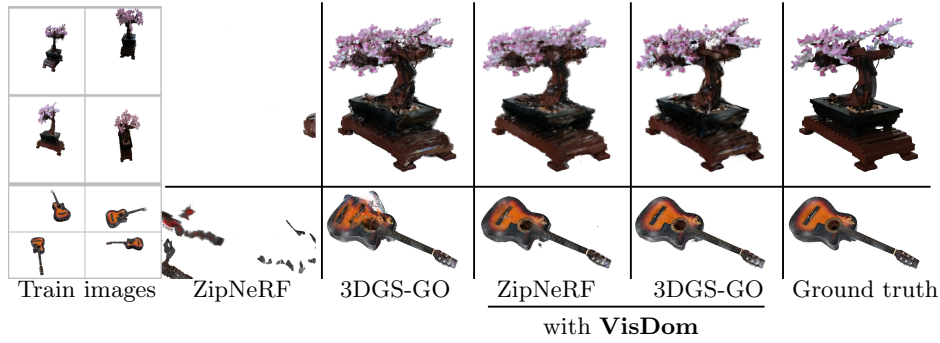
\begin{figure*}
    \centering
    \begin{overpic}[width=1.0\linewidth, tics=5,]
    {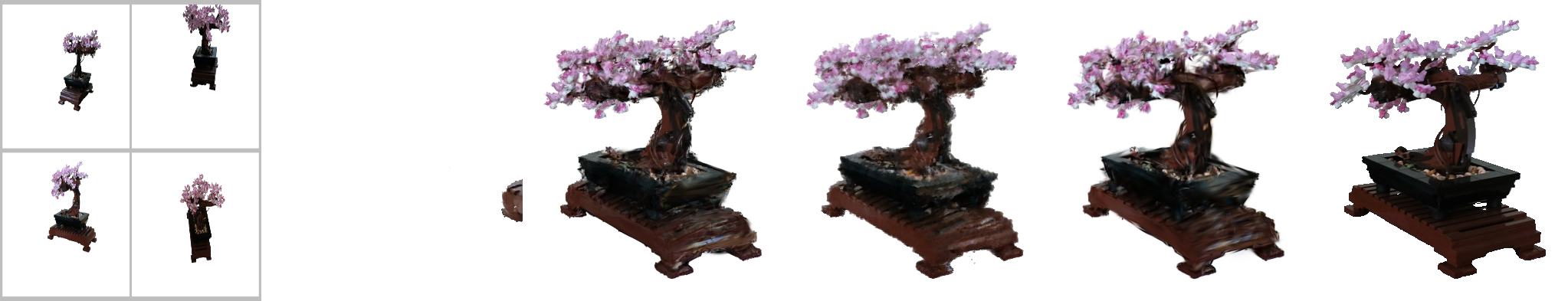}
    \end{overpic}
    \begin{overpic}[width=1.0\linewidth, tics=5,]
    {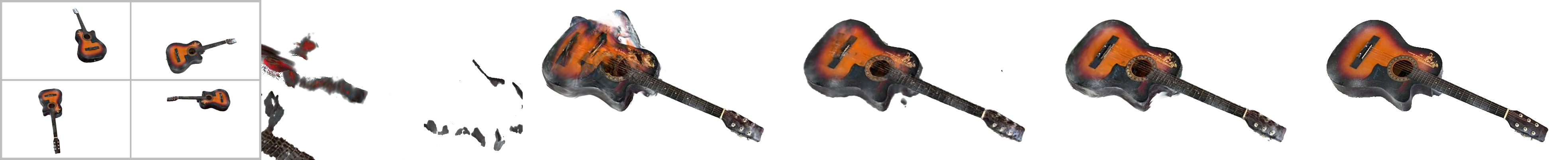}
    \put(1,-2){\small Train images}
    \put(20,-2){\small ZipNeRF}
    \put(36,-2){\small 3DGS-GO}
    \put(53,-2){\small ZipNeRF}
    \put(70,-2){\small 3DGS-GO}
    \put(17,10){\linethickness{0.25mm}\color{black}\line(1,0){83}}%
    \put(52,-3){\linethickness{0.25mm}\color{black}\line(1,0){30}}%
    \put(58,-6){\small with \textbf{VisDom}}
    \put(85,-2){\small Ground truth}
    \put(34,0){\linethickness{0.25mm}\color{black}\line(0,1){30}}%
    \put(50.5,0){\linethickness{0.25mm}\color{black}\line(0,1){30}}%
    \put(67.5,0){\linethickness{0.25mm}\color{black}\line(0,1){30}}%
    \put(84,0){\linethickness{0.25mm}\color{black}\line(0,1){30}}
    \end{overpic}
    \vspace{2mm}
    \caption{Our learning-free geometric constraint, derived purely from silhouettes, enables reconstruction from as few as 4 images. Given 4 inputs (left), ZipNeRF~\cite{barron2023zipnerf} and 3DGS-GO~\cite{yang2024gaussianobject} (cols 1-2) struggle without our constraint; adding VisDom (cols 3-4) recovers high-quality reconstructions with zero additionally learned parameters.\vspace{-5mm}}
    \label{fig:teaser}
\end{figure*}
\begin{abstract}
Sparse novel view synthesis (NVS) remains challenging due to the ambiguity of recovering 3D geometry from few input views. While NeRF- and Gaussian Splatting (GS)-based methods perform well with dense supervision, they often overfit in sparse settings, producing floating artifacts and inconsistent geometry. Silhouette consistency is commonly used as a regularizer, but it remains insufficient, as silhouette-consistent regions can extend beyond the true object geometry. We introduce VisDom, a learning-free geometric constraint that augments classical carving-based visual hull reconstruction by enforcing a minimum multi-view visibility requirement. Specifically, we define a visible domain as the subset of 3D space observed by at least $K$ views and use it as an additional filtering criterion on top of standard silhouette-based reconstruction. This provides a stronger spatial prior in sparse-view settings. We integrate VisDom into both implicit (NeRF) and explicit (GS) pipelines by restricting volumetric sampling and guiding Gaussian placement during optimization. Experiments on three challenging datasets show consistent improvements in sparse-view NVS, enabling high-quality object-centric reconstruction from as few as four input images. Our method is domain-agnostic, requires only silhouettes, and introduces no learned parameters, making it a simple complement to existing approaches. Applying VisDom on top of GaussianObject further improves performance on Omni3D and MipNeRF360, while 3DGS-GO + VD matches or surpasses GaussianObject at 22 $\times$ lower training cost.

\end{abstract}

\section{Introduction}
Inferring missing information from sparse observations is a long-standing challenge in computer vision. A
particularly prominent instance of this problem is \emph{novel view synthesis} (NVS), where the goal is to
generate unseen views of a scene given only a set of input images. The remarkable progress in neural radiance
fields (NeRFs)~\cite{mildenhall2020nerf,mueller2022instant,barron2023zipnerf} and 3D Gaussian Splatting
(GS)~\cite{kerbl3Dgaussians,yu2024mipgs} has established NVS as a central research problem, enabling
photorealistic rendering and efficient training pipelines. While these methods achieve impressive results when
dense multi-view input is available, their performance deteriorates significantly in sparse-view scenarios,
where recovering the underlying 3D structure becomes ill-posed.
The sparse-view regime is practically critical as dense capture is infeasible in casual 3D capture, robotics,
and AR/VR, yet the reconstruction task is fundamentally ill-posed. NeRFs allocate density arbitrarily along rays to fit training colors,
while GS relies on COLMAP~\cite{schonberger2016colmap} initialization that itself fails under sparse input.
Prior regularization strategies introduce additional complexity through learned priors~\cite{Niemeyer2021Regnerf,kim2022infonerf,wang2023sparsenerf}, depth constraints~\cite{deng2022dsnerf}, or diffusion guidance~\cite{wynn2023diffusionerf}, and may depend on domain-specific assumptions or require additional domain-specific training data that limit generalization.

Among the available signals, object silhouettes offer an attractive compromise: they are simple to extract
from off-the-shelf segmentation models such as SAM~\cite{kirillov2023sam}, and they directly encode both
occupancy and free space. Multi-view silhouettes have a long history in 3D reconstruction, yet their role in
modern sparse-view NVS has remained limited. While silhouette consistency is a natural regularizer, it can be
counter-productive in extreme sparsity: we observe that adding a silhouette loss to ZipNeRF actually
\emph{degrades} PSNR at 4 views relative to vanilla training, because the constraint is too weak to carve out
the large ambiguous region admitted by so few silhouettes, see ``traditional'' visual hull in ~\cref{fig:visualspace}. In other words, silhouettes alone cannot fully
resolve the depth uncertainty introduced by sparse input views; a stronger geometric prior is needed.

This motivates us to revisit the classical idea of lifting silhouettes into 3D space. We target the object-centric, outside-in $360^\circ$ capture setting, where multi-view silhouettes are most informative. Our key intuition is
that even in sparse settings, enforcing a coarse 3D geometric support can significantly reduce reconstruction ambiguity. 

To this end, we introduce \textbf{VisDom}, a novel \emph{visible domain} constraint that augments the
visual hull. Unlike traditional visual hulls, which directly intersect multi-view silhouettes and often
overestimate geometry, our visible domain constraint defines the region that is jointly observed in at least
$K$ views. This modification yields a more reliable 3D geometric prior in the sparse regime as demonstrated
in~\cref{fig:visualspace}. We integrate VisDom into NeRF's volumetric rendering by restricting ray sampling to the
constrained domain; for Gaussian splatting, we enforce the constraint by regulating point placement and
allocation during optimization, demonstrating consistent visual gains, see~\cref{fig:teaser}.

Crucially, VisDom introduces no learned parameters and imposes no domain-specific assumptions, making it a natural complement to existing learned-prior methods. Applying VisDom on top of GaussianObject~\cite{yang2024gaussianobject} further improves results on object-centric datasets, while the learning-free 3DGS-GO + VD independently surpasses GO on out-of-distribution domains such as human subjects, where generative priors degrade.

Our contributions can be summarized as follows:
\begin{itemize}
    \item We introduce \textbf{VisDom}, a learning-free visible-domain constraint that augments classical carving-based visual hull reconstruction with an additional multi-view visibility filtering stage, requiring only a 2-second pre-processing step and zero learned parameters.
    \item We design principled integration strategies for both volumetric (NeRF) and explicit (3DGS) rendering
    paradigms to benefit from multi-view silhouette geometry with our visible-domain constraint.
    \item We empirically demonstrate consistent improvements across five diverse NVS frameworks on three
    challenging real-world datasets, recovering general-purpose methods that completely fail
    without our constraint - \textit{from $\sim$12 dB to $\sim$25 dB PSNR at 4 views} - and advancing the state of the
    art among sparse methods, with 3DGS-GO + VD matching sparse-view methods in quality while training up to
    \textit{$22\times$ faster}.
\end{itemize}

\section{Related Work}
\noindent \textbf{Sparse Multi-View Reconstruction.}
Most sparse NVS approaches rely on learned shape priors~\cite{yu2020pixelnerf,Niemeyer2020dvr,chibane2021srf}
or depth supervision~\cite{deng2022dsnerf,yu2022monosdf,li2024dngaussian} to reconstruct from a single image or a few views.
RegNeRF~\cite{Niemeyer2021Regnerf} and DiffusionNeRF~\cite{wynn2023diffusionerf} use learned priors to
regularize image patches for front-facing scenes, but are limited to small viewpoint deviations rather than
full 360$^\circ$ object capture. GaussianObject~\cite{yang2024gaussianobject} applies
ControlNet~\cite{zhang2023adding} diffusion to refine an initial Gaussian splatting reconstruction;
FreeNeRF~\cite{yang2023freenerf} controls positional encoding frequencies; ZeroRF~\cite{shi2023zerorf} and
SplatFields~\cite{SplatFields} use implicit neural networks as a prior. Recent Sparse2DGS~\cite{wu2025sparse2dgs} addresses sparse-view surface reconstruction by initializing Gaussians from depth estimates provided by a learned MVS network. VisDom takes a different approach:
it imposes a \textit{learning-free} geometric constraint derived solely from multi-view silhouettes, adding no learned
parameters and remaining domain-agnostic. Rather than competing with learned-prior methods, VisDom
complements them --- effective standalone and further boosting learned-prior methods when combined.

\noindent \textbf{Visual Hull for Novel View Synthesis.}
Visual hull~\cite{laurentini1994visualhull}  is a classical concept of a maximum shape consistent with a set
of multi-view silhouettes. One can estimate a reasonable mesh from a set of silhouettes easily obtained from
off-the-shelf segmentation networks. Only a few works advocate the use of visual hulls.
VaxNeRF~\cite{kondo2021vaxnerf} found its use in significantly reduced training time in a dense regime, as the
visual hull volume allows for the pruning of network queries by dropping space samples outside. However, our
work primarily focuses on the benefits of a visible domain-constrained visual hull for sparse reconstructions.
Moreover, models trained with our visible domain constraint implicitly learn the scene's geometry, while
VaxNeRF relies on the visual hull, \textit{even during inference}, to produce a faithful reconstruction.
The visual hull has been used in recent Gaussian splatting-based
methods~\cite{yang2024gaussianobject,SplatFields} to bypass COLMAP limitations in estimating a reliable
initialization for sparse camera configurations. However, our method tightly integrates our visible
domain-constrained visual hull in the training process. %

\section{Background}
\subsection{Visual Hull and Space Carving}
\label{subsec:vhull}

A visual hull is a geometric shape derived from multiple 2D silhouettes of an object captured from different viewpoints. Formally, it is the \emph{intersection} of the viewing cones back-projected from each silhouette, i.e., the maximal 3D volume whose projection is consistent with \emph{all} observed silhouettes~\cite{laurentini1994visualhull}. As more views are added, this intersection tightens and more closely approximates the observed object. Pure shape-from-silhouette cannot recover concavities that never appear on a silhouette boundary, although certain concave shapes are exactly recoverable~\cite{franco2003exact}; classical space carving~\cite{kutulakos2000carve} therefore augments silhouette carving with photo-consistency to recover such concavities~\cite{sinha2005multi}. We emphasize that hard voxel carving of this kind is a long-standing tool in volumetric reconstruction (e.g., tomography): our contribution is not the hardness of the constraint, but the multi-view \emph{visibility} filtering that makes the carved hull reliable in the sparse-view NVS regime (\cref{subsec:vhull_sparse}). While there exists a large corpus of Shape-from-Silhouettes (SFS) works from the 3D reconstruction community~\cite{baumgart1974geometricmodeling,laurentini1994visualhull,franco2003exact}, we focus on the practical voxel-based formulation commonly adopted in NeRF applications~\cite{kondo2021vaxnerf,yang2024gaussianobject}, inspired by space carving~\cite{kutulakos2000carve}.

\begin{figure*}[ht]
    \centering
    \begin{overpic}[width=\linewidth,tics=5,]
        {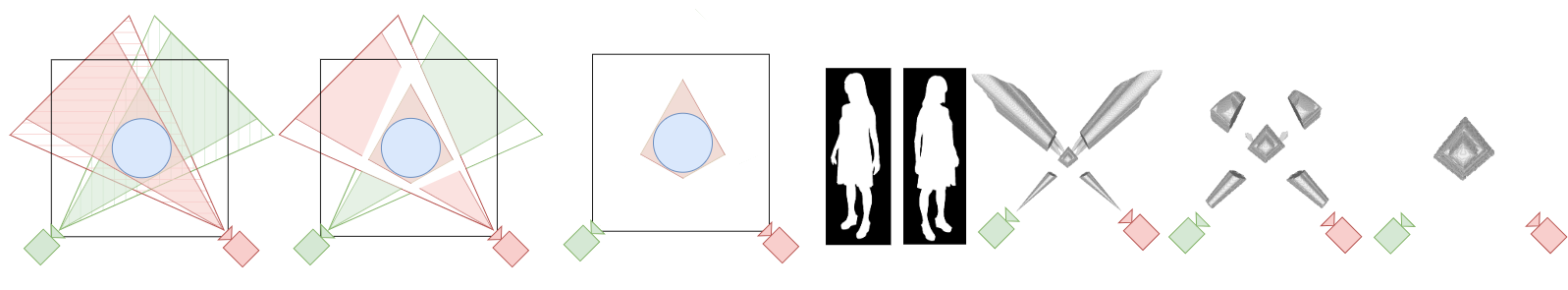}
        \put(15,-7){ \small Visual Hull in 2D}
        \put(2,0){ \tiny Capture Setup}
        \put(21,0){ \tiny traditional}
        \put(23, -2){ \tiny vhull}

        \put(39,0){ \tiny \textbf{VisDom}}
         \put(51.5,-2){\linethickness{0.2mm}\color{black}\line(0,1){22}}%

        \put(60,-7){ \small Visual Hull in 3D (top view)}

        \put(51.5,0){ \tiny Given posed}
        \put(52,-2){ \tiny silhouettes}

        \put(63,0){ \tiny traditional}
        \put(65, -2){ \tiny vhull}

        \put(75,0){ \tiny unit sphere}
        \put(76 ,-2){ \tiny clipping}
        \put(77 ,-4){\tiny ~\cite{kondo2021vaxnerf,yang2024gaussianobject}}
        \put(89,0){ \tiny \textbf{VisDom}}
    \end{overpic}
    \vspace{4mm}
    \caption{Visual hull reconstruction in 2D (left part) and top-view in 3D (right part). Left part: Given
    $2$ silhouettes of an object (left), the traditional visual hull (middle) is the \emph{intersection} of the
    two silhouette cones; with only two views this intersection is inherently loose and retains a large
    ambiguous region. In this sparse-view regime, much of the retained volume consists of voxels supported by
    only a \emph{single} camera --- they are silhouette-consistent yet lack any cross-view evidence for their
    occupancy. Our visible-domain constraint removes exactly this space, keeping only voxels jointly observed by
    at least $K$ views (right). Right part: Given $2$ posed silhouettes (left), the traditional visual hull
    retains elongated single-camera regions along the viewing cones (middle left). Some
    methods~\cite{kondo2021vaxnerf,yang2024gaussianobject} cull regions outside a unit sphere, assuming a
    centered object (middle right). Our constraint instead prunes regions not co-observed by the required number
    of views (right), without any assumption on object placement.
    }
    \label{fig:visualspace}
\end{figure*}
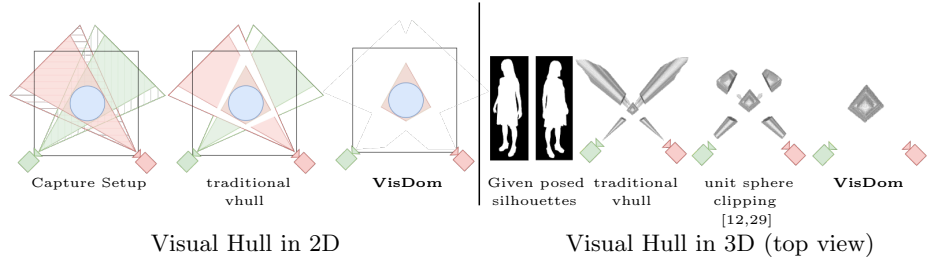

To compute a shape from silhouettes, we initialize two voxel grids representing occupancy and visibility. We project voxel centers into the posed images. For each view, voxels whose projections fall within the silhouette receive an occupancy vote, while those whose projections fall within the image bounds receive a visibility vote. A voxel is marked as occupied if its occupancy votes exceed $X\%$ (e.g., $ 95\%$) of its visibility votes, accounting for inaccuracies in the provided masks. Finally, marching cubes~\cite{mc1997Lorensen} is applied to extract a mesh from the occupied voxel grid. A visualization of a reconstructed visual hull is shown in~\cref{fig:visualspace}.
\section{Our Method}
We aim to reconstruct an object given its sparse posed images and masks. Towards that, we introduce a
constraint called the ``visible domain'' and utilize it to obtain a reliable visual hull
(\cref{subsec:vhull_sparse}).
We use the estimated shape to enforce the visible domain constraint in NeRFs trained with silhouette loss
(\cref{sec:nerfwithvhull}) and 3DGS (\cref{subsec:visdom_gaussplat}).
Before diving into our contributions, we provide an in-depth analysis of the limitations of the silhouette
constraints alone for sparse reconstruction in~\cref{subsec:problem_statement}.
\subsection{Limitations of Silhouette-only Constraints for Sparse Reconstruction}
\label{subsec:problem_statement}
NeRF and 3DGS face significant challenges in reconstructing faithful 360$^{\circ}$ scenes from sparse
multi-view images. Both approaches allocate densities (NeRF) or place Gaussians (GS) based on training views
without global geometric regularization, often resulting in floaters and structural inconsistencies.
Since these artifacts contribute to minimizing the training loss, they cannot be removed in post-processing;
an effective sparse strategy must instead control \textit{where} density is allocated.
Multi-view silhouette constraints provide a partial solution: methods such as~\cite{yariv2020idr,sun2021nelf}
use silhouette losses to carve out space in which density can be assigned, conceptually resembling visual hull
carving (\cref{subsec:vhull}). However, in extremely sparse settings this is insufficient. As shown
in~\cref{fig:visualspace}, the visual hull from few views can be very large, and without explicit multi-view
correspondences, densities are placed wherever the rendering loss is minimal — not necessarily at the true
object location. We therefore focus on the 3D regions observed by \textit{at least $K$} cameras — the
``visible domain'' — to tighten the visual hull and eliminate ambiguous space before optimization begins.
\begin{figure}[ht]
\centering
\begin{minipage}[t]{0.55\linewidth}
    \centering
    \begin{overpic}[width=\linewidth,tics=5,]
        {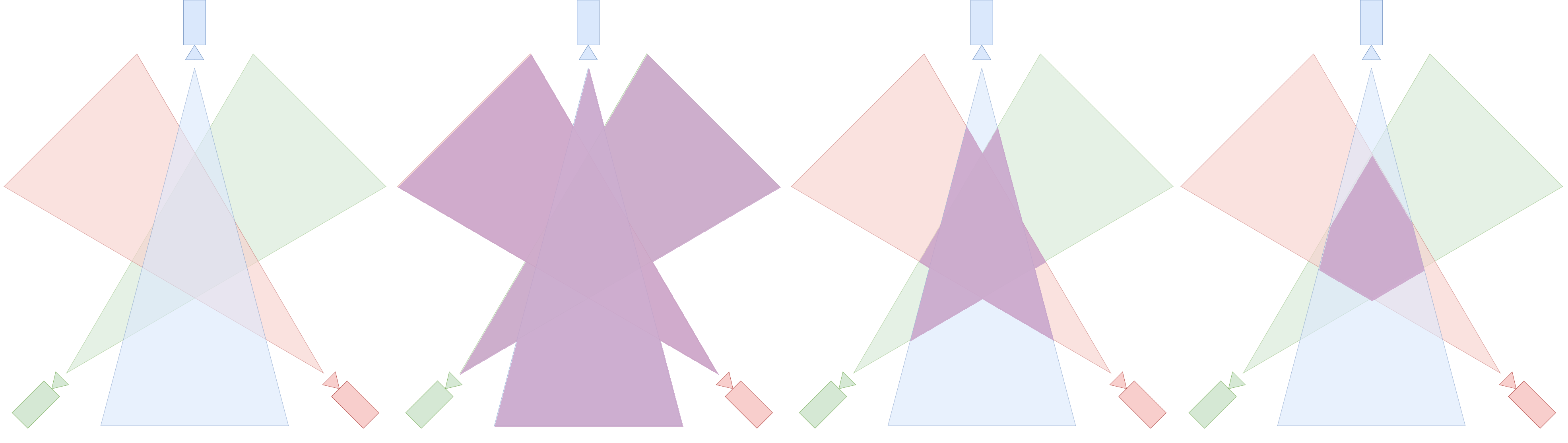}
         \put(2,-8){\tiny Camera setup}%
         \put(25,-8){\linethickness{0.2mm}\color{black}\line(0,1){38}}%
         \put(33,-4){\tiny $K=1$}%
         \put(58,-4){\tiny $K=2$}%
         \put(81,-4){\tiny $K=3$}%
         \put(50,-8){\tiny Visible domains}%
    \end{overpic}
    \caption{A 2D example of the visible domain (purple) for different values of $K$ in a $3$-camera setup.
    Larger $K$ values narrow the covered region, leading to more precise shapes.}
    \label{fig:viewingdomain}
\end{minipage}%
\hfill%
\begin{minipage}[t]{0.42\linewidth}
    \centering
    \begin{overpic}[width=\linewidth,tics=5,]
        {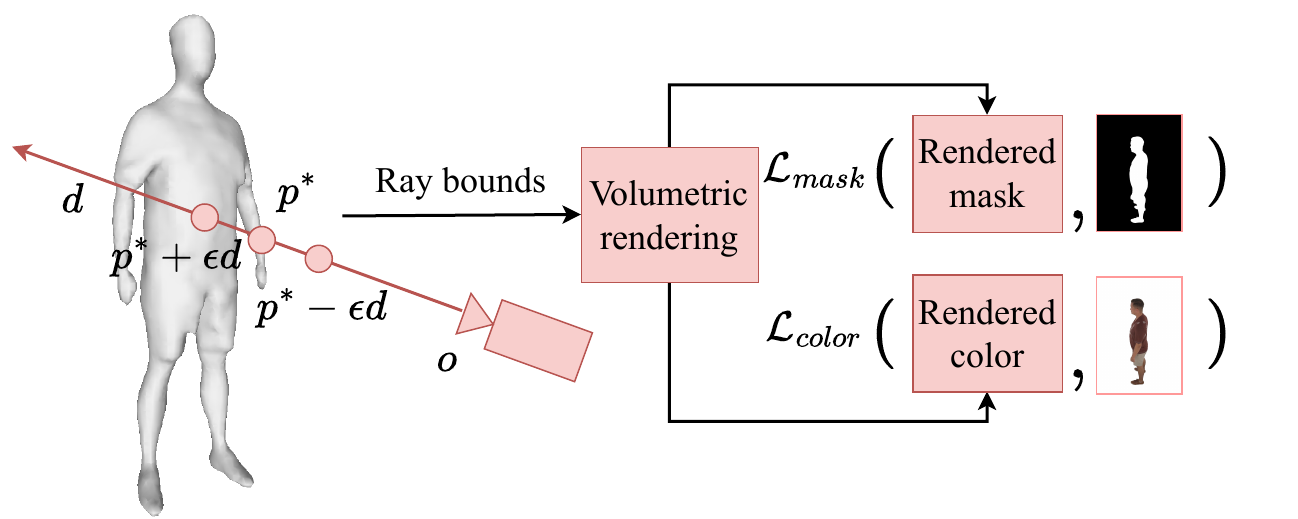}
         \put(9,-3){\tiny vhull}%
         \put(1,-7){\tiny with VisDom}%
         \put(50,-3){\tiny Constrained}%
         \put(41,-7){\tiny volumetric rendering}%
    \end{overpic}
    \caption{Our VisDom visual hull (left) confines ray sampling bounds during NeRF volumetric rendering
    (right), enforcing a geometric constraint on the allocated densities.}
    \label{fig:pipeline}
\end{minipage}
\end{figure}
\subsection{Visual Hull from Sparse Views}
\label{subsec:vhull_sparse}

The carving-based visual hull approach (\cref{subsec:vhull}) removes voxels inconsistent with silhouettes, but can still retain regions that are weakly constrained in sparse-view settings. In particular, voxels observed by only a single camera may remain silhouette-consistent despite being geometrically unreliable. These regions often correspond to unobserved space behind the subject and are a primary source of undesired density allocations in sparse reconstruction settings (\cref{fig:visualspace}).

To address this issue, we introduce the concept of the visible domain, defined as the subset of 3D space observed by at least $K$ cameras. For $K=1$, the visible domain corresponds to the union of all viewing frusta. As $K$ increases, the visible domain becomes progressively more compact, retaining only regions jointly observed across multiple views, as illustrated in~\cref{fig:viewingdomain}.

Importantly, the visible domain constraint is applied in addition to the standard carving-based visual hull formulation, rather than replacing it. We first compute the visual hull using the occupancy and visibility voting procedure described in~\cref{subsec:vhull}. During the final occupancy filtering step, in addition to enforcing that occupancy votes are at least $X\%$ of the visibility votes, we also require that the visibility count is at least $K$. In practice, this suppresses voxels supported by only a small number of views, which are common in sparse-view capture setups.

A common alternative to suppress unwanted voxel allocations is to assume that the object is centered in a canonical coordinate system and clip regions outside a unit sphere, as done in VaxNeRF~\cite{kondo2021vaxnerf} and GaussianObject~\cite{yang2024gaussianobject}. However, such assumptions limit applicability to controlled-capture scenarios where object centering can be guaranteed. In contrast, the visible domain constraint is entirely data-driven and independent of assumptions about object position or scale, making it broadly applicable across different capture configurations.
\subsection{VisDom Constraint for Sparse NVS}
We propose to regularize reconstruction methods using a robust version of the visual hull. For readers
unfamiliar with the underlying representations, we provide concise overviews of
NeRFs~\cite{mildenhall2020nerf} and 3D Gaussian Splatting (3DGS)~\cite{kerbl3Dgaussians} in the supplementary
material. 

\subsubsection{Neural Radiance Fields.}
\label{sec:nerfwithvhull}
We consider posed RGB images and masks as inputs. We first compute a visual hull imposing both silhouette and
visible domain constraints as discussed in~\ref{subsec:vhull_sparse}.
We restrain each ray's sampling range $[t_n, t_f]$ by intersecting it with the sparse visual hull. Let
$\mathbf{p}^*$ be the point of intersection between a camera ray $\mathbf{r} = (\mathbf{o}, \mathbf{d})$,
parameterized with an origin $\mathbf{o}$ and a direction $\mathbf{d}$, and the sparse visual hull. Then, the
ray's sampling bounds become $[t_n = \mathbf{p}^* - \epsilon\mathbf{d}, t_f = \mathbf{p}^* +
\epsilon\mathbf{d}]$, where $\epsilon$ is a small scalar, as illustrated in~\cref{fig:pipeline}. Importantly,
$\epsilon$ is deliberately non-zero: it defines a band \emph{around} the hull rather than pinning samples to
its surface, giving the network freedom to place the true surface slightly inside the hull -- e.g., for mild
concavities where the hull encloses but does not coincide with the object -- while still excluding the large
ambiguous free space. In practice we use a generous $\epsilon{=}0.2$ (\cref{subsec:implementation}), so the
constraint bounds sampling without forcing the surface onto the hull. Our
modification is a simple yet effective solution for sparse reconstruction that can be easily integrated into
the volumetric rendering of any NeRF pipeline.\\
In this work, we train two popular NeRF-based methods, ZipNeRF~\cite{barron2023zipnerf} and
InstantNGP~\cite{mueller2022instant}, with our proposed constraint, along with silhouette losses, while
maintaining the training procedures of the respective methods. The optimized objective can be defined as
\begin{align}\label{eq:our_loss_nerf}
    \mathcal{L}_\text{ours} = \mathcal{L}_\text{base} & + \lambda_\text{1} \mathcal{L}_\text{mask}\,,
\end{align}
where $\mathcal{L}_\text{base}$ is the training loss of the base method (ZipNeRF or InstantNGP),
$\mathcal{L}_\text{mask}$ is the binary cross entropy loss between the given and the rendered masks, and
$\lambda$ is the weight of the silhouette loss.\\
\subsubsection{Gaussian Splatting}
\label{subsec:visdom_gaussplat}
Gaussian splatting~\cite{kerbl3Dgaussians} (3DGS) is a 3D scene representation based on 3D Gaussian learned
per scene. As the representation is explicit, bounding the near and far planes, as with NeRFs, is no longer an
option. Consequently, unlike the hard sampling restriction we impose on NeRF, the visible-domain constraint enters 3DGS as a \emph{soft} penalty. We use our visible domain-constrained visual hull to regularize 3DGS in two ways. Firstly, we
initialize the 3DGS reconstruction, similar to existing works~\cite{SplatFields,yang2024gaussianobject}, with
the visual hull obtained with the algorithm as described in ~\cref{subsec:vhull_sparse}. Secondly, we
interpolate between the camera poses of the training set and enforce a visibility constraint on the new views.
As discussed in~\cref{subsec:problem_statement}, we can identify 3D regions where Gaussians are not expected
due to the visual hull. Therefore, we enforce the mask loss on unoccupied pixels in the unseen views, refining
the reconstruction by adjusting the Gaussians’ opacities.
This ensures that the reconstructed Gaussians are inside our visual domain-constrained visual hull. The
proposed loss looks as follows
\begin{align}\label{eq:our_loss_gauss}
    \mathcal{L}_\text{ours} = \mathcal{L}_\text{base} + \lambda_\text{2} \mathcal{L}_\text{mask}  + \lambda_\text{3} \mathcal{L}_\text{i}\,, 
\end{align}
where $\mathcal{L}_\text{i} = -(1-M_i) \log(1-\hat{M}_i)$ penalizes Gaussians that appear opaque in regions outside the visual hull when rendered from interpolated camera views. Here, masks $M_i$ are obtained by rendering our visual hull from these novel views.
$\mathcal{L}_\text{base}$ and $\mathcal{L}_\text{mask}$ are the
loss of the base method (3DGS-GO) and the binary cross-entropy loss between the given and the rendered masks, respectively.

\begin{table*}[!ht]
\centering
\setlength\tabcolsep{2pt}
\scalebox{0.76}{
\begin{tabular}{c|c|c|c|c|c|c|c|c|c|c|c|c}
\multirow{2}{*}{Dataset} & \multirow{2}{*}{Cams} & \multirow{2}{*}{VaxNeRF} & \multirow{2}{*}{ZeroRF} & Splat & \multirow{2}{*}{FSGS} & \multirow{2}{*}{CoR-GS} & \multirow{2}{*}{GO} & \multicolumn{5}{c}{\textbf{with our VisDom constraint}} \\
 & & & & Fields & & & & INGP & ZipNeRF & 3DGS-GO & CoR-GS & GO\\ \hline

\multicolumn{2}{c|}{Train time} & $2$ hrs & $2$ hrs & $1$ hr & $14$ min & \cellcolor{tabsecond}$10$ min & $45$ min & $20$ min & $40$ min & \cellcolor{tabfirst}$2$ min & \cellcolor{tabsecond}$10$ min & $45$ min \\ \hline

& & \multicolumn{11}{c}{PSNR $\uparrow$} \\ \cline{2-13}

\multirow{4}{*}{\rotatebox[origin=c]{90}{Mip360}}
 & $4$    & 19.19 & 14.17 & 22.24 & 23.38 & 24.04 & 24.02 & 22.15 & 24.10 & 24.06 & \cellcolor{tabfirst}24.64 & \cellcolor{tabsecond}24.16 \\
 & $6$    & 21.28 & 24.14 & 24.57 & 26.17 & 25.81 & 26.23 & 24.14 & 25.80 & \cellcolor{tabsecond}26.72 & \cellcolor{tabfirst}27.35 & 26.50 \\
 & $9$    & 22.32 & 27.78 & 26.58 & 28.16 & \cellcolor{tabsecond}28.58 & 27.94 & 25.43 & 28.06 & 28.45 & \cellcolor{tabfirst}29.32 & 28.14 \\
 & Mean   & 20.93 & 22.03 & 24.46 & 25.91 & 26.14 & 26.06 & 23.91 & 25.99 & \cellcolor{tabsecond}26.41 & \cellcolor{tabfirst}27.10 & 26.27 \\ \hline

\multirow{4}{*}{\rotatebox[origin=c]{90}{Omni3D}}
 & $4$    & 18.33 & 27.78 & 28.49 & 27.31 & 28.97 & \cellcolor{tabsecond}30.37 & 27.44 & 29.49 & 30.32 & 29.82 & \cellcolor{tabfirst}30.71 \\
 & $6$    & 19.52 & 31.94 & 32.05 & 29.74 & 32.51 & 33.26 & 29.67 & 32.28 & \cellcolor{tabfirst}33.37 & 32.87 & \cellcolor{tabsecond}33.29 \\
 & $9$    & 20.76 & 32.93 & 34.66 & 33.46 & 34.94 & 35.56 & 31.06 & 35.21 & \cellcolor{tabfirst}35.69 & 35.56 & \cellcolor{tabsecond}35.59 \\
 & Mean   & 19.54 & 30.88 & 31.73 & 30.17 & 32.14 & 33.06 & 29.39 & 32.33 & \cellcolor{tabsecond}33.12 & 32.75 & \cellcolor{tabfirst}33.20 \\ \hline

\multirow{4}{*}{\rotatebox[origin=c]{90}{ActorsHQ}}
 & $5$    & 19.25 & 25.13 & 22.16 & 24.44 & 24.94 & 24.91 & 23.53 & 24.55 & \cellcolor{tabfirst}25.69 & \cellcolor{tabsecond}25.27 & 24.85 \\
 & $8$    & 21.35 & 26.47 & 24.67 & 26.48 & 26.93 & 26.98 & 23.35 & 26.72 & \cellcolor{tabfirst}27.99 & \cellcolor{tabsecond}27.06 & 26.87 \\
 & $12$   & 23.04 & 27.59 & 26.90 & 27.96 & 28.21 & 28.17 & 25.67 & \cellcolor{tabsecond}28.61 & \cellcolor{tabfirst}29.13 & 28.22 & 28.10 \\
 & Mean   & 21.21 & 26.40 & 24.58 & 26.29 & 26.69 & 26.69 & 24.18 & 26.63 & \cellcolor{tabfirst}27.60 & \cellcolor{tabsecond}26.85 & 26.61 \\ \hline

& & \multicolumn{11}{c}{SSIM $\uparrow$} \\ \cline{2-13}

\multirow{4}{*}{\rotatebox[origin=c]{90}{Mip360}}
 & $4$    & 0.8332 & 0.8188 & 0.9107 & 0.9258 & \cellcolor{tabsecond}0.9280 & 0.9270 & 0.9054 & 0.9171 & 0.9243 & \cellcolor{tabfirst}0.9323 & 0.9271 \\
 & $6$    & 0.8736 & 0.9211 & 0.9328 & 0.9438 & 0.9404 & 0.9460 & 0.9258 & 0.9452 & 0.9461 & \cellcolor{tabfirst}0.9499 & \cellcolor{tabsecond}0.9468 \\
 & $9$    & 0.8958 & 0.9460 & 0.9488 & 0.9581 & 0.9609 & 0.9600 & 0.9316 & \cellcolor{tabsecond}0.9620 & 0.9596 & \cellcolor{tabfirst}0.9632 & 0.9605 \\
 & Mean   & 0.8675 & 0.8953 & 0.9308 & 0.9426 & 0.9431 & 0.9443 & 0.9209 & 0.9414 & 0.9433 & \cellcolor{tabfirst}0.9484 & \cellcolor{tabsecond}0.9448 \\ \hline

\multirow{4}{*}{\rotatebox[origin=c]{90}{Omni3D}}
 & $4$    & 0.8884 & 0.9615 & 0.9660 & 0.9617 & 0.9660 & \cellcolor{tabsecond}0.9731 & 0.9547 & 0.9694 & 0.9705 & 0.9702 & \cellcolor{tabfirst}0.9735 \\
 & $6$    & 0.9056 & 0.9731 & 0.9761 & 0.9713 & 0.9766 & \cellcolor{tabfirst}0.9817 & 0.9620 & 0.9793 & 0.9802 & 0.9785 & \cellcolor{tabsecond}0.9813 \\
 & $9$    & 0.9106 & 0.9747 & 0.9824 & 0.9813 & 0.9831 & \cellcolor{tabfirst}0.9869 & 0.9647 & 0.9865 & 0.9863 & 0.9850 & \cellcolor{tabsecond}0.9868 \\
 & Mean   & 0.9016 & 0.9698 & 0.9748 & 0.9714 & 0.9752 & \cellcolor{tabsecond}0.9806 & 0.9605 & 0.9784 & 0.9790 & 0.9779 & \cellcolor{tabfirst}0.9806 \\ \hline

\multirow{4}{*}{\rotatebox[origin=c]{90}{ActorsHQ}}
 & $5$    & 0.8171 & 0.9118 & 0.8780 & 0.9185 & \cellcolor{tabsecond}0.9206 & 0.9082 & 0.8713 & 0.9125 & 0.9162 & \cellcolor{tabfirst}0.9277 & 0.9104 \\
 & $8$    & 0.7768 & 0.9210 & 0.8912 & \cellcolor{tabsecond}0.9295 & 0.9254 & 0.9250 & 0.8602 & \cellcolor{tabfirst}0.9322 & 0.9294 & 0.9260 & 0.9253 \\
 & $12$   & 0.7612 & 0.9264 & 0.9065 & \cellcolor{tabsecond}0.9377 & 0.9350 & 0.9354 & 0.8750 & \cellcolor{tabfirst}0.9393 & 0.9375 & 0.9351 & 0.9338 \\
 & Mean   & 0.7850 & 0.9197 & 0.8919 & \cellcolor{tabsecond}0.9286 & 0.9270 & 0.9229 & 0.8689 & 0.9280 & 0.9277 & \cellcolor{tabfirst}0.9296 & 0.9231
\end{tabular}
}
\caption{Quantitative comparison (PSNR$\uparrow$ top block, SSIM$\uparrow$ bottom block) of VisDom applied to
sparse-specific methods against sparse NVS baselines. Colors indicate \colorbox{tabfirst}{the best} and
\colorbox{tabsecond}{the second best} per row. CoR-GS+VD achieves the best results on MipNeRF360; 3DGS+VD leads
on Omni3D (per view) and ActorsHQ. GO+VD achieves the best Omni3D mean, but underperforms on ActorsHQ due to
the domain gap for humans in GO's pre-trained diffusion model. SSIM shows consistent trends: CoR-GS+VD and
GO+VD attain the best or second-best SSIM in most settings, and VisDom methods remain competitive on ActorsHQ,
where SSIM saturates and differences are within the third decimal.
\vspace{-3mm}}
\label{tab:quantitativesota}
\end{table*}

\begin{table*}[ht]
\centering
\setlength\tabcolsep{2pt}
\scalebox{0.92}{
\begin{tabular}{l|c|ccc|ccc|cc}
\multirow{2}{*}{Dataset} & \multirow{2}{*}{Views}
& \multicolumn{3}{c|}{INGP~\cite{mueller2022instant}}
& \multicolumn{3}{c|}{ZipNeRF~\cite{barron2023zipnerf}}
& \multicolumn{2}{c}{3DGS-GO~\cite{yang2024gaussianobject}} \\
\cmidrule(lr){3-5} \cmidrule(lr){6-8} \cmidrule(lr){9-10}
& & van. & +mask & +VD & van. & +mask & +VD & van.* & +VD \\
\midrule
& & \multicolumn{8}{c}{PSNR$\uparrow$} \\
\cmidrule(lr){3-10}
\multirow{3}{*}{\rotatebox[origin=c]{90}{\small Mip360}}
 & 4 & 13.67 & 13.73 & \textbf{22.15} & 12.44 & 11.95 & \textbf{24.10} & 23.61 & \textbf{24.06} \\
 & 6 & 14.73 & 17.36 & \textbf{24.14} & 11.89 & 12.78 & \textbf{25.80} & 26.30 & \textbf{26.72} \\
 & 9 & 16.04 & 20.74 & \textbf{25.43} & 14.85 & 19.87 & \textbf{28.06} & 27.93 & \textbf{28.45} \\
\midrule
\multirow{3}{*}{\rotatebox[origin=c]{90}{\small Omni3D}}
 & 4 & 18.88 & 18.16 & \textbf{27.44} & 15.04 & 15.02 & \textbf{29.49} & 29.80 & \textbf{30.32} \\
 & 6 & 19.95 & 22.01 & \textbf{29.67} & 16.61 & 20.00 & \textbf{32.28} & 33.09 & \textbf{33.37} \\
 & 9 & 20.86 & 22.31 & \textbf{31.06} & 23.16 & 25.30 & \textbf{35.21} & 35.49 & \textbf{35.69} \\
\midrule
\multirow{3}{*}{\rotatebox[origin=c]{90}{\small ActHQ}}
 & 5  & 12.01 & 13.87 & \textbf{23.53} & 10.85 & 10.33 & \textbf{24.55} & 25.12 & \textbf{25.69} \\
 & 8  & 13.86 & 14.96 & \textbf{23.35} & 11.43 & 10.76 & \textbf{26.72} & 27.44 & \textbf{27.99} \\
 & 12 & 20.98 & 24.86 & \textbf{25.67} & 11.32 & 28.38 & \textbf{28.61} & 28.80 & \textbf{29.13} \\
\midrule
& & \multicolumn{8}{c}{SSIM$\uparrow$} \\
\cmidrule(lr){3-10}
\multirow{3}{*}{\rotatebox[origin=c]{90}{\small Mip360}}
 & 4 & 0.7831 & 0.7557 & \textbf{0.9054} & 0.8400 & 0.8227 & \textbf{0.9171} & 0.9216 & \textbf{0.9243} \\
 & 6 & 0.7969 & 0.8083 & \textbf{0.9258} & 0.8118 & 0.8103 & \textbf{0.9452} & 0.9442 & \textbf{0.9461} \\
 & 9 & 0.8187 & 0.8678 & \textbf{0.9316} & 0.8049 & 0.8812 & \textbf{0.9620} & 0.9580 & \textbf{0.9596} \\
\midrule
\multirow{3}{*}{\rotatebox[origin=c]{90}{\small Omni3D}}
 & 4 & 0.8820 & 0.8671 & \textbf{0.9547} & 0.8939 & 0.8803 & \textbf{0.9694} & 0.9687 & \textbf{0.9705} \\
 & 6 & 0.8924 & 0.9046 & \textbf{0.9620} & 0.8993 & 0.9253 & \textbf{0.9793} & 0.9794 & \textbf{0.9802} \\
 & 9 & 0.8827 & 0.9128 & \textbf{0.9647} & 0.9334 & 0.9515 & \textbf{0.9865} & 0.9857 & \textbf{0.9863} \\
\midrule
\multirow{3}{*}{\rotatebox[origin=c]{90}{\small ActHQ}}
 & 5  & 0.7568 & 0.7378 & \textbf{0.8713} & 0.8264 & 0.7988 & \textbf{0.9125} & 0.9014 & \textbf{0.9162} \\
 & 8  & 0.7482 & 0.7648 & \textbf{0.8602} & 0.8356 & 0.8063 & \textbf{0.9322} & 0.9202 & \textbf{0.9294} \\
 & 12 & 0.8626 & \textbf{0.9118} & 0.8750 & 0.8229 & \textbf{0.9432} & 0.9393 & 0.9324 & \textbf{0.9375} \\
\end{tabular}
}
\caption{Quantitative evaluation (PSNR$\uparrow$ top block, SSIM$\uparrow$ bottom block) of VisDom on
general-purpose NeRF and 3DGS methods. Best variant per method group is in \textbf{bold}. \emph{van.*} - the
initialization stage of GaussianObject~\cite{yang2024gaussianobject} (3DGS with depth and silhouette priors).
VisDom consistently improves all methods and substantially outperforms silhouette-only training; the only exceptions are the densest ActorsHQ setting (12 views), where silhouette supervision alone (+mask) already
suffices in SSIM.\vspace{-1mm}}
\label{tab:sparserec}
\end{table*}
\begin{figure*}[ht]
\centering
    \begin{overpic}[width=0.85\linewidth,tics=5,]
        {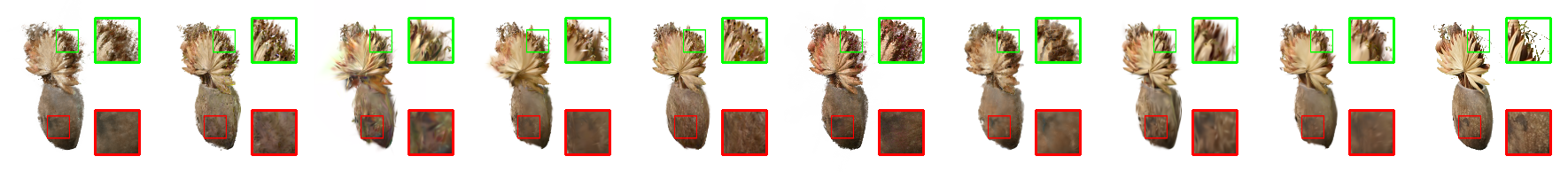}
        \put(-4,3){\rotatebox{90}{{\tiny $4$ cams}}}
    \end{overpic}
    \begin{overpic}[width=0.85\linewidth,tics=5,]
        {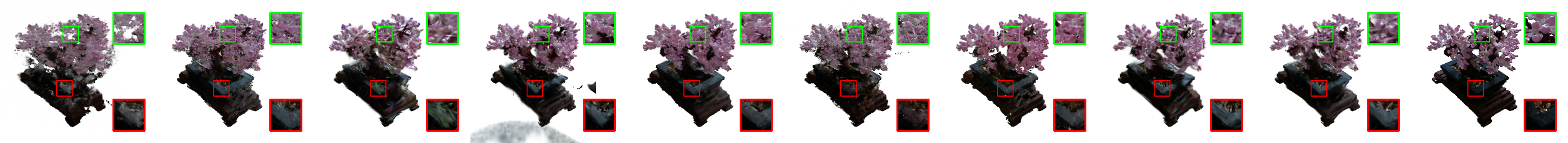}
        \put(-4,3){\rotatebox{90}{{\tiny $6$ cams}}}
        \put(-8,0){\rotatebox{90}{{\small Mip360}}}
    \end{overpic}
    \begin{overpic}[width=0.85\linewidth,tics=5,]
        {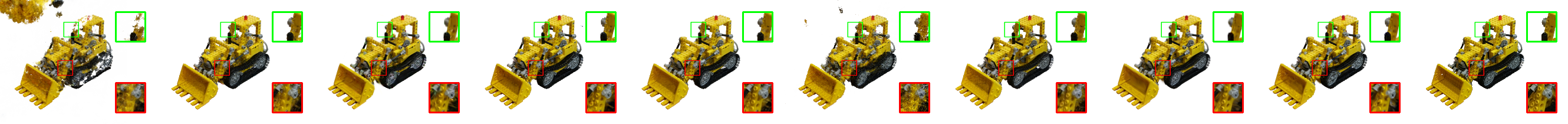}
        \put(-4,2){\rotatebox{90}{{\tiny $9$ cams}}}
    \end{overpic}
    \begin{overpic}[width=0.85\linewidth,tics=5,]
        {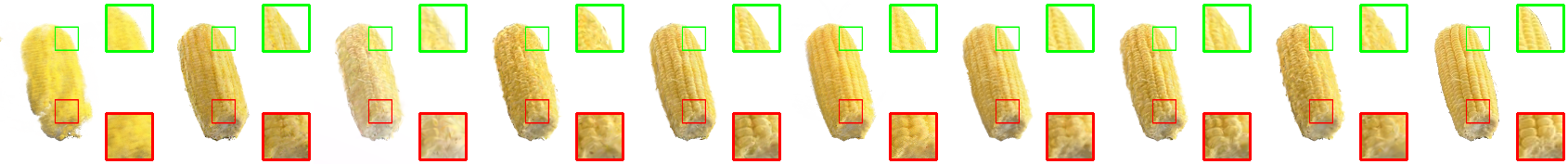}
        \put(-4,3){\rotatebox{90}{{\tiny $4$ cams}}}
    \end{overpic}
    \begin{overpic}[width=0.85\linewidth,tics=5,]
        {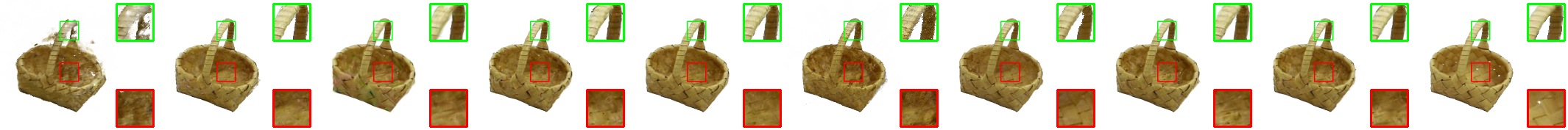}
        \put(-4,2){\rotatebox{90}{{\tiny $6$ cams}}}
        \put(-8,1){\rotatebox{90}{{\small Omni3D}}}
    \end{overpic}
    \begin{overpic}[width=0.85\linewidth,tics=5,]
        {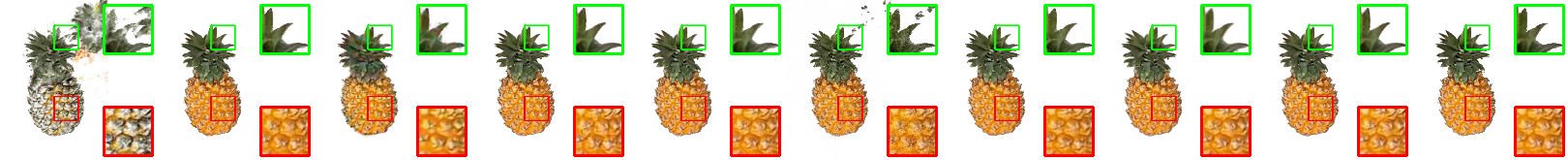}
        \put(-4,2.5){\rotatebox{90}{{\tiny $9$ cams}}}
    \end{overpic}
        \begin{overpic}[width=0.85\linewidth,tics=5,]
        {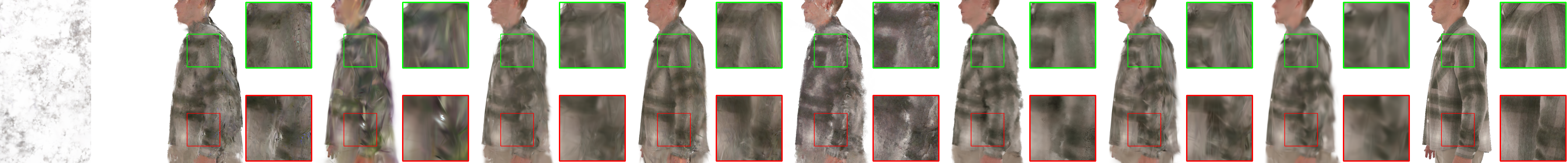}
        \put(-4,2){\rotatebox{90}{{\tiny $5$ cams}}}
    \end{overpic}
    \begin{overpic}[width=0.85\linewidth,tics=5,]
        {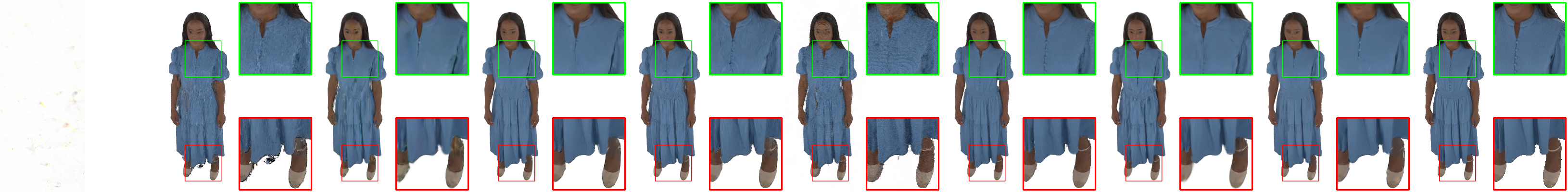}
        \put(-8,3){\rotatebox{90}{{\small ActorsHQ}}}
        \put(-4,3){\rotatebox{90}{{\tiny $8$ cams}}}
    \end{overpic}
    \begin{overpic}[width=0.85\linewidth,tics=5,]
        {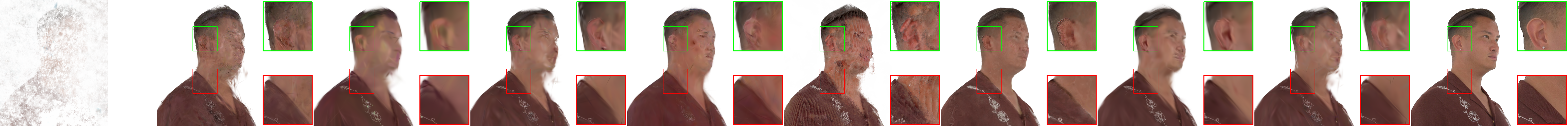}
        \put(-4,0){\rotatebox{90}{{\tiny $12$ cams}}}

         \put(0,-1.8){\tiny VaxNeRF}
         \put(10,-1.8){\tiny SplatFields}
         \put(22,-1.8){\tiny FSGS}
         \put(30,-1.8){\tiny CoR-GS}
         \put(43,-1.8){\tiny GO}
         \put(52,-1.8){\tiny INGP}
         \put(60,-1.8){\tiny ZNeRF}
         \put(70,-1.8){\tiny 3DGS-GO}
         \put(81,-1.8){\tiny CoR-GS}
         \put(91,-1.8){\tiny Ground}
         \put(92.5,-3){\tiny truth}

        \put(50,-2.5){\linethickness{0.2mm}\color{black}\line(1,0){40}}%
         \put(65,-4){\tiny + VisDom}
    \end{overpic}
    \vspace{2mm}
    \caption{Qualitative comparison of methods trained with our constraint and other NVS approaches in sparse
    settings on three diverse, challenging datasets, MipNeRF360, Omni3D, and ActorsHQ. Our constraint improves
    state-of-the-art sparse reconstruction methods such as CoR-GS and 3DGS-GO, and enables existing NeRF methods
    to perform comparably to sparse-view baselines. Best seen zoomed-in.}
    \label{fig:qualitativesota}
\end{figure*}
\section{Experiments}
We evaluate VisDom on ActorsHQ~\cite{isik2023humanrf}, MipNeRF360~\cite{barron2022mipnerf360} (Mip360), and
Omni3D~\cite{brazil2023omni3d}, spanning humans and general objects in 360$^\circ$ object-centric settings.
\Cref{subsec:compare_sota} benchmarks all VisDom-enhanced models against sparse-specific baselines.
\Cref{subsec:improvingnerf} then ablates VisDom on general-purpose methods —
ZipNeRF~\cite{barron2023zipnerf}, Instant-NGP~\cite{mueller2022instant}, and 3DGS-GO — isolating its effect
from sparse-specific inductive biases. Due to page limit, further evaluation details, including LPIPS, are provided in the supplementary.
\subsection{Implementation}
\label{subsec:implementation}
Training and evaluation of all models were conducted on an RTX 4090 GPU. We utilized original hyperparameters and settings for the existing baselines. For the methods with our constraint, we enforce the silhouette loss with $\lambda_\text{1} = 0.1$ (\cref{eq:our_loss_nerf}) and visual
hull-based ray-object intersection sampling for ZipNeRF and InstantNGP as discussed
in~\cref{sec:nerfwithvhull}. For 3DGS-based regularization (\cref{eq:our_loss_gauss}) we set $\lambda_\text{2}
= 0.1$ and $\lambda_\text{3} = 0.01$.
\subsubsection{Visual hull reconstruction} 
If ground truth masks are provided in the dataset (e.g.
ActorsHQ~\cite{isik2023humanrf}), they drive our reconstruction of a visual hull. For the other datasets, such as Omni3D~\cite{brazil2023omni3d} and MipNeRF360~\cite{barron2022mipnerf360}, we use masks released
by~\cite{yang2024gaussianobject}. Using the silhouettes of each set of training cameras, we use the algorithm
proposed in~\cref{subsec:vhull_sparse} to reconstruct the visual hull. We set the minimum number of observing
cameras $K=3$ and run marching cubes to obtain the mesh of the reconstructed shape from the voxel grid. We
constrain the volumetric rendering using the visual hull obtained from silhouettes by rendering the mesh with
the given poses to obtain the intersection distance. We set $\epsilon=0.2$ for the near and far plane
distances.

Our computation of the visual hull takes about $2$ seconds. Moreover, the VisDom constraint imposes no additional training overhead, since the visual hull is computed once during pre-processing. Due to space constraints, we provide further evaluations of the visual hull with our constraint against other variants (traditional and unit-sphere) in the supplementary material.

\subsection{Comparison against Sparse NVS Methods}
\label{subsec:compare_sota}
We demonstrate VisDom's plug-and-play nature on sparse-specific methods; \cref{subsec:improvingnerf} then
ablates its isolated effect on general-purpose methods. We apply the visible-domain constraint to
CoR-GS~\cite{zhang2024corgs}, a method designed for sparse object reconstruction, and to GaussianObject
(GO)~\cite{yang2024gaussianobject}, which extends 3DGS-GO with ControlNet-based diffusion refinement. We
benchmark all five VisDom-enhanced models against VaxNeRF~\cite{kondo2021vaxnerf},
ZeroRF~\cite{shi2023zerorf}, SplatFields~\cite{SplatFields}, FSGS~\cite{zhu2024fsgs}, CoR-GS, and GO. For a
fair comparison with VaxNeRF, we use Instant-NGP as the base model and do not use the visual hull during
inference.

Results in~\cref{tab:quantitativesota} demonstrate that VisDom improves every method it is applied to. CoR-GS+VD achieves the highest mean PSNR on MipNeRF360, outperforming all baselines,
including methods that rely on generative priors or neural regularizers. 3DGS-GO+VD leads on Omni3D (per-view)
and ActorsHQ. GO+VD achieves the best mean on Omni3D but underperforms on ActorsHQ, consistent with the domain
gap for human subjects in GO's pre-trained diffusion model — an effect that does not affect our geometry-only
constraint. Crucially, 3DGS-GO+VD trains in just \textit{2 minutes per scene} — $10\times$ faster than Instant-NGP+VD and
$22\times$ faster than GO — while remaining competitive or superior in reconstruction quality. 
This demonstrates that VisDom's geometric constraint delivers high-quality results with zero additional learned parameters, while remaining complementary to methods that incorporate generative priors.

Qualitative comparisons in~\cref{fig:qualitativesota,fig:qualitativesparse} confirm that applying VisDom constraint results in fewer floaters, sharper
detail, and more consistent geometry. We provide additional qualitative results in the supplementary material.

\subsection{VisDom on Existing NeRF and 3DGS Models} \label{subsec:improvingnerf}
We evaluate two NeRF methods, ZipNeRF~\cite{barron2023zipnerf} and Instant-NGP~\cite{mueller2022instant}, and
3DGS initialized using the GaussianObject~\cite{yang2024gaussianobject} pipeline (3DGS-GO). These are
general-purpose methods that carry no sparse-specific inductive biases, enabling a clean ablation of VisDom's
effect. For NeRF methods we evaluate three variants: (i) vanilla, (ii) silhouette loss only (+mask), and (iii)
VisDom (+VD). For 3DGS-GO we compare vanilla initialization against +VD. Results are in~\cref{tab:sparserec}.

As shown in~\cref{tab:sparserec}, adding silhouette supervision (+mask) sometimes \emph{degrades} performance
relative to the vanilla baseline - for example, ZipNeRF+mask drops from 12.44 to 11.95 dB at 4 views on
Mip360, and similarly on ActorsHQ. This occurs because sparse silhouettes define a visual hull that is far too
large to constrain the reconstruction: the model is penalized for densities outside the hull, but the hull
itself spans an enormous, ambiguous volume. VisDom resolves this by restricting reconstruction to the jointly
visible region, turning these failures into competitive results. ZipNeRF+VD at 4 views reaches 24.10 dB on
Mip360 ($\sim$90\% PSNR gain), recovering a method that was previously unusable at this sparsity level.
Instant-NGP+VD achieves a $\sim$60\% boost. As camera count increases, silhouettes become more discriminative,
and the gap between +mask and +VD narrows, confirming that VisDom is especially critical under extreme
sparsity. 3DGS-GO already benefits from depth and silhouette priors, yet VisDom still delivers consistent
gains, demonstrating complementarity with existing structural regularizers.
\begin{figure*}[ht]
    \centering
    \begin{overpic}[width=1\linewidth,tics=5, ]
        {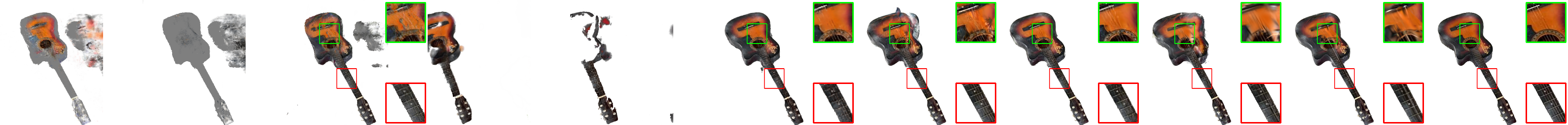}
         \put(-2,1){\rotatebox{90}{\tiny $4$ cams}}
    \end{overpic}
    \begin{overpic}[width=1\linewidth,tics=5, ]
        {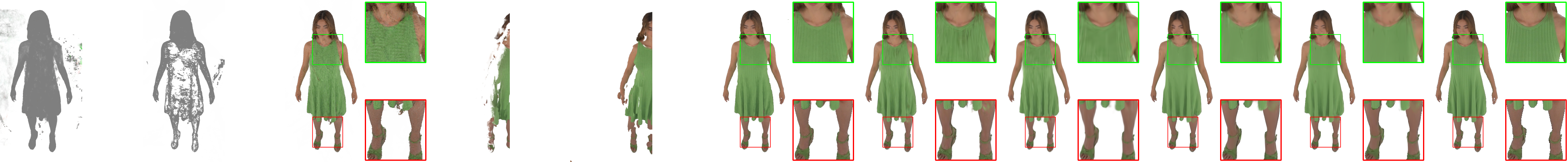}
         \put(-2,2){\rotatebox{90}{\tiny $8$ cams}}
    \end{overpic}
    \begin{overpic}[width=1\linewidth,tics=5,]
        {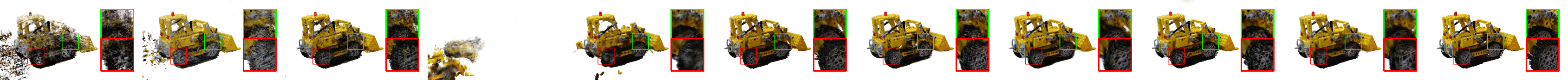}
         \put(-2,-1){\rotatebox{90}{\tiny $9$ cams}}

         \put(2,-1.5){\tiny vanilla}
         \put(10,-1.5){\tiny +mask}
         \put(19,-1.5){\tiny +VisDom}
         \put(7,-5){\small Instant-NGP}

         \put(30,-1.5){\tiny vanilla}
         \put(38,-1.5){\tiny +mask}
         \put(46,-1.5){\tiny +VisDom}
         \put(35,-5){\small ZipNeRF}

         \put(56,-1.5){\tiny vanilla}
         \put(64,-1.5){\tiny +VisDom}
         \put(58,-5){\small 3DGS-GO}

         \put(74,-1.5){\tiny vanilla}
         \put(82,-1.5){\tiny +VisDom}
         \put(77,-5){\small CoR-GS}

         \put(91,-3){\small Ground}
         \put(92.5,-5){\small truth}

        \put(27,0){\linethickness{0.25mm}\color{black}\line(0,1){25}}%
        \put(55,0){\linethickness{0.25mm}\color{black}\line(0,1){25}}%
        \put(73,0){\linethickness{0.25mm}\color{black}\line(0,1){25}}%
        \put(91,0){\linethickness{0.25mm}\color{black}\line(0,1){25}}%
    \end{overpic}
    \vspace{1mm}
    \caption{Qualitative evaluation of our constraints on NeRF and 3DGS methods. Our constraints show dramatic
    improvement in rendering quality for NeRF-based methods. Due to our visible domain regularization, we
    observe fewer artifacts, especially at object boundaries in 3DGS-based reconstructions. Best seen zoomed-in.}
    \label{fig:qualitativesparse}
    \vspace{2mm}
\end{figure*}
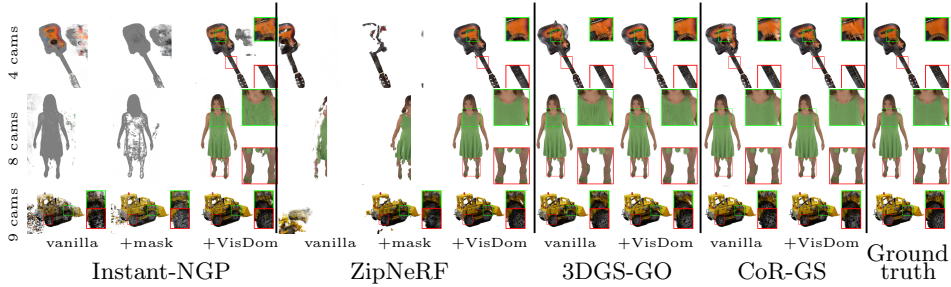%
Further, we provide qualitative results in~\cref{fig:qualitativesparse}, which visually demonstrate the
insufficiency of silhouettes for NeRF methods in sparse settings and the effectiveness of our constraints for
both NeRF- and 3DGS-based models. In particular, NeRF-based approaches often fail to achieve multi-view
consistent reconstructions when prioritizing photometric consistency across wide-baseline training views. Our
visible-domain constraint confines the reconstruction volume, ensuring that density is allocated only within
the maximal space the object can occupy. This regularization proves especially beneficial for 3DGS methods in
extremely sparse settings, effectively removing ghost-like floaters from the reconstruction, as demonstrated
by the guitar example in~\cref{fig:qualitativesparse}.

\subsection{Ablation: Minimum Number of Observing Cameras $K$}
\label{subsec:ablation_k}
\begin{table}[h!]
\centering
\setlength\tabcolsep{3.5pt}
\scalebox{0.90}{
\begin{tabular}{c|cccc|cccc}
\multirow{2}{*}{Views}
  & \multicolumn{4}{c|}{ZipNeRF+VD}
  & \multicolumn{4}{c}{3DGS-GO+VD} \\
\cmidrule(lr){2-5}\cmidrule(lr){6-9}
  & $K{=}1$ & $K{=}2$ & $K{=}3$ & $K{=}4$
  & $K{=}1$ & $K{=}2$ & $K{=}3$ & $K{=}4$ \\
\midrule
  & \multicolumn{8}{c}{PSNR$\uparrow$} \\
\cmidrule(lr){2-9}
$4$ & 13.55 & 22.71 & 23.26 & \textbf{23.79} & 13.23 & 23.98 & \textbf{24.06} & 24.00 \\
$6$ & 16.21 & 21.47 & \textbf{25.80} & 22.04 & 18.23 & \textbf{27.17} & 26.72 & 26.56 \\
$9$ & 18.66 & 28.06 & 28.06 & \textbf{28.71} & 24.16 & \textbf{29.06} & 28.45 & 28.54 \\
\midrule
Mean & 16.14 & 24.08 & \textbf{25.71} & 24.84 & 18.54 & \textbf{26.74} & 26.41 & 26.36 \\
\end{tabular}
}
\caption{Ablation of the minimum number of observing cameras~$K$ on MipNeRF360 (PSNR$\uparrow$). Bold indicates the best result per row
and method. We prioritize a balance between overly permissive and aggressive volume carving in our selection of $K{=}3$. \vspace{-1mm}}
\label{tab:ablation_k}
\end{table}

The visual hull is obtained by voxel carving: a voxel is retained only if it is observed by at
least $K$ cameras (\cref{subsec:vhull_sparse}). 
We ablate $K \in \{1, 2, 3, 4\}$ on MipNeRF360 for both ZipNeRF and 3DGS-GO in~\cref{tab:ablation_k}.
When any single silhouette suffices to retain a voxel, the reconstructed hull is nearly
unconstrained, collapsing to a near-vanilla setting. Without cross-view occupancy evidence, rays still traverse a large, ambiguous
region. As a result, NeRF density accumulates spuriously, and 3DGS floaters remain unsuppressed.
Requiring agreement from at least two cameras eliminates the majority of ambiguous space and
immediately recovers most of the performance gap. ZipNeRF improves further from $K{=}2$ to
$K{=}3$, benefiting from the tighter hull that confines
NeRF sampling to the jointly visible object region; $K{=}3$ achieves the best mean for ZipNeRF. For 3DGS-GO, $K{=}2$ peaks higher than $K{=}3$ on average, but the looser hull it produces still admits
surface regions visible from only a single camera, leaving residual floaters that degrade a
subset of test views. Note that for this experiment, we do not apply strongly connected components-based culling after computing the visual hull.

Although $K{=}4$ marginally outperforms $K{=}3$ at several view counts for ZipNeRF, $K{=}3$ achieves the best mean across all view counts ($25.71$ vs. $24.84$ dB) and is most robust at the hardest 4-view setting, where high $K$ risks over-carving surface regions visible from only a few cameras.

\subsection{Robustness to Mask Quality}
\label{subsec:maskrobust}
Since VisDom is derived from foreground silhouettes, we assess its sensitivity to mask imperfections by
dilating the input masks with a square structuring element of radius $r$ before hull reconstruction. On
MipNeRF360, ZipNeRF+VD is stable and even improves slightly under moderate dilation (e.g., $+0.58\,$dB at
$r{=}3$ for 4 views), showing that a small safety margin compensates for multi-view mask inconsistencies and imperfections rather than harming the hull. 
We report $r{=}0$
throughout the main results to avoid dataset-specific tuning and provide the full sweep
in the supplementary.

\section{Limitations}
\label{sec:limits}
VisDom is designed for object-centric, outside-in $360^\circ$ capture, and its benefit stems from cross-view co-visibility. The gains are largest when input views are well distributed around the object and diminish when views are clustered from similar directions, where the visible domain barely tightens the hull, and depth- or diffusion-based supervision may be preferable. The constraint further assumes reasonably accurate camera poses and foreground silhouettes. Retaining robustness to pose inaccuracies and enabling pose-free operation remain part of future work.
Despite delivering consistent improvements across nearly all tested models and datasets, in scenes with strong inter-view lighting variation (e.g., MipNeRF360), Visdom and geometric constraints alone are insufficient without a generative prior, as reflected in the ZipNeRF+VD vs.\ GO gap in~\cref{fig:qualitativesota}. 
\section{Conclusion}
We presented VisDom, a learning-free geometric constraint that tightens the classical visual hull by
restricting reconstruction to the region jointly visible in at least $K$ views. A key finding is that silhouette supervision alone is insufficient at extreme sparsity: with only a few views, the set of silhouette-consistent configurations remains large, so the loss regularizes renderings without resolving geometry. VisDom resolves this by enforcing K-view co-visibility, removing ambiguous volume before optimization begins. VisDom resolves this by enforcing $K$-view co-visibility, removing
ambiguous volume that silhouettes cannot resolve. The constraint adds only a 2-second preprocessing step and
zero learned parameters, and integrates with any NeRF or 3DGS pipeline via a single modification.

Across five reconstruction frameworks and three real-world datasets, VisDom consistently improves quality,
enabling general-purpose methods that otherwise fail (up to $90\%$ PSNR gain at 4 views) and advancing sparse
models (CoR-GS + VD on MipNeRF360; 3DGS-GO + VD on ActorsHQ) without any additional learned
parameters.

\bibliographystyle{splncs04}
\bibliography{main}

\clearpage
\appendix
\section*{Supplementary Material ---\\ VisDom: Sparse Novel View Synthesis with Visible Domain Constraint}

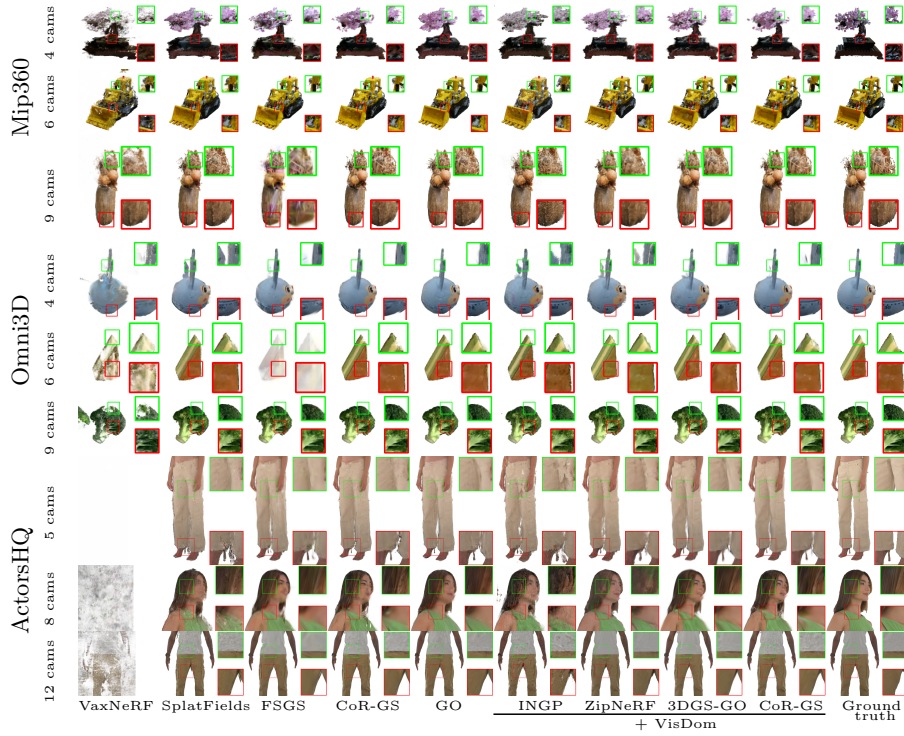
\begin{figure*}[ht]
\centering
    \begin{overpic}[width=0.9\linewidth,tics=5,]
        {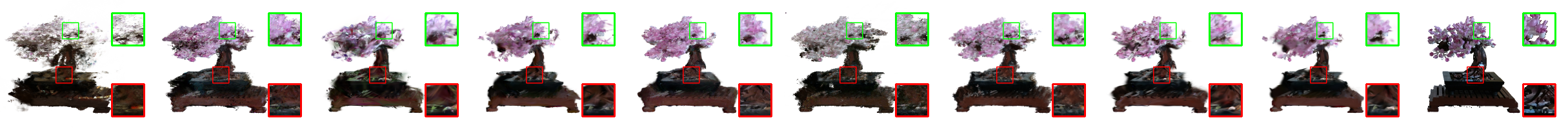}
        \put(-4,1){\rotatebox{90}{{\tiny $4$ cams}}}
    \end{overpic}
    \begin{overpic}[width=0.9\linewidth,tics=5,]
        {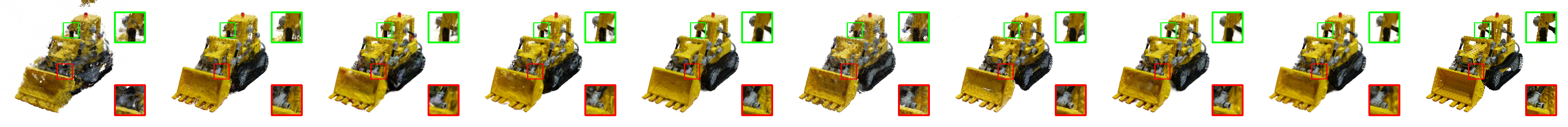}
        \put(-4,1){\rotatebox{90}{{\tiny $6$ cams}}}
        \put(-8,0){\rotatebox{90}{{\small Mip360}}}
    \end{overpic}
    \begin{overpic}[width=0.9\linewidth,tics=5,]
        {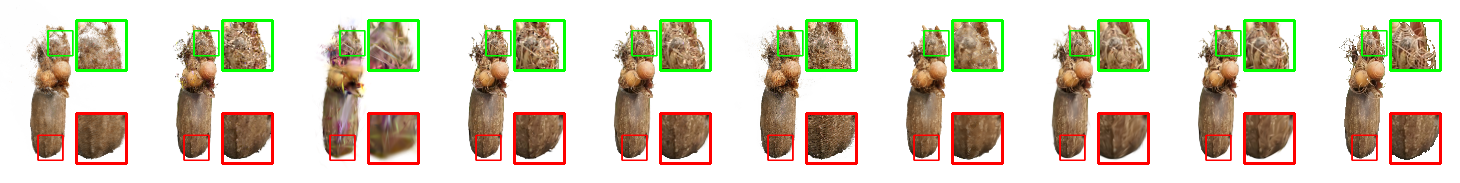}
        \put(-4,2){\rotatebox{90}{{\tiny $9$ cams}}}
    \end{overpic}
    \begin{overpic}[width=0.9\linewidth,tics=5,]
        {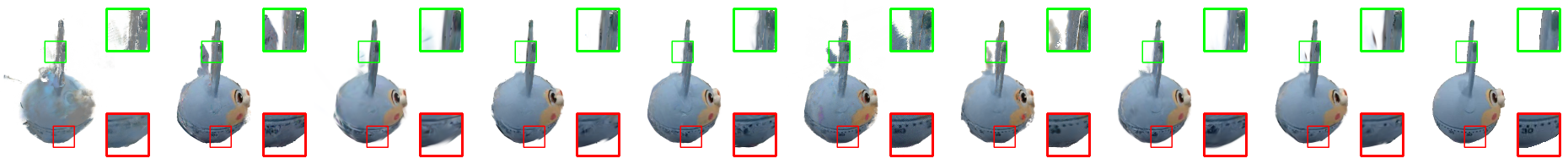}
        \put(-4,2){\rotatebox{90}{{\tiny $4$ cams}}}
    \end{overpic}
    \begin{overpic}[width=0.9\linewidth,tics=5,]
        {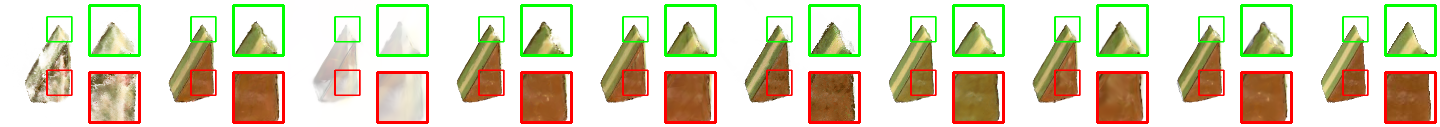}
        \put(-4,1){\rotatebox{90}{{\tiny $6$ cams}}}
        \put(-8,1){\rotatebox{90}{{\small Omni3D}}}
    \end{overpic}
    \begin{overpic}[width=0.9\linewidth,tics=5,]
        {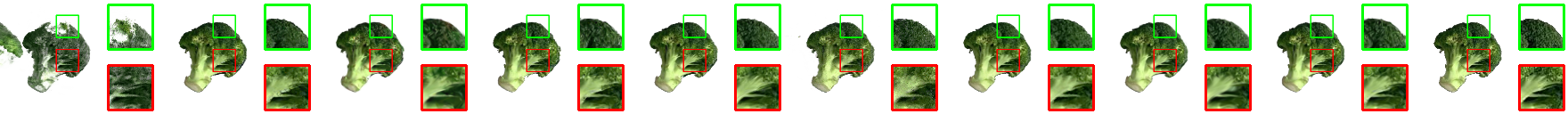}
        \put(-4,0){\rotatebox{90}{{\tiny $9$ cams}}}
    \end{overpic}
        \begin{overpic}[width=0.9\linewidth,tics=5,]
        {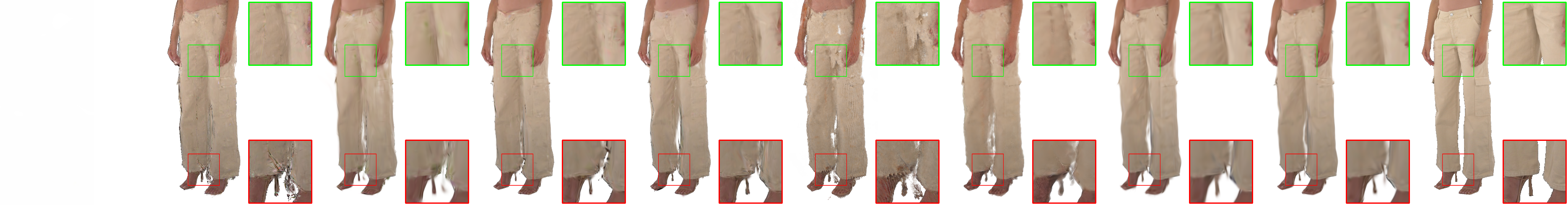}
        \put(-4,3){\rotatebox{90}{{\tiny $5$ cams}}}
    \end{overpic}
    \begin{overpic}[width=0.9\linewidth,tics=5,]
        {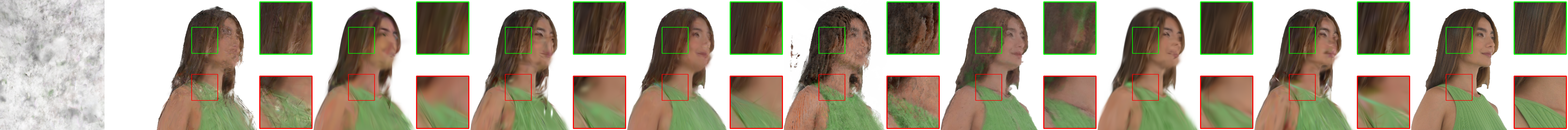}
        \put(-8,0){\rotatebox{90}{{\small ActorsHQ}}}
        \put(-4,1){\rotatebox{90}{{\tiny $8$ cams}}}
    \end{overpic}
    \begin{overpic}[width=0.9\linewidth,tics=5,]
        {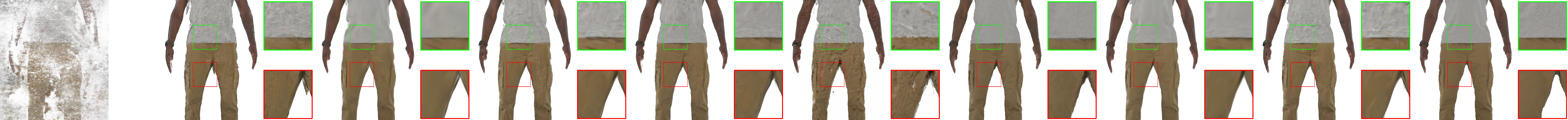}
        \put(-4,-1){\rotatebox{90}{{\tiny $12$ cams}}}

         \put(0,-1.8){\tiny VaxNeRF}
         \put(10,-1.8){\tiny SplatFields}
         \put(22,-1.8){\tiny FSGS}
         \put(31,-1.8){\tiny CoR-GS}
         \put(43,-1.8){\tiny GO}
         \put(53,-1.8){\tiny INGP}
         \put(61,-1.8){\tiny ZipNeRF}
         \put(71,-1.8){\tiny 3DGS-GO}
         \put(82,-1.8){\tiny CoR-GS}
         \put(92,-1.8){\tiny Ground}
         \put(93.5,-3){\tiny truth}
         
        \put(50,-2.3){\linethickness{0.2mm}\color{black}\line(1,0){40}}%
         \put(67,-4){\tiny + VisDom}
    \end{overpic}

    \vspace{3mm}
    \caption{Additional qualitative results on novel view synthesis from sparse views ($4$, $6$, and $9$) from
    diverse, challenging datasets --- MipNeRF360, Omni3D, and ActorsHQ. As can be seen, our geometric
    constraint helps ZipNeRF and Instant-NGP achieve high-quality reconstructions from a sparse set of images.
    Further, our constraint also helps improve the quality of reconstructions of SOTA 3DGS-based methods, such
    as a robust version of 3DGS-GO~\cite{yang2024gaussianobject} and CoR-GS~\cite{zhang2024corgs}.}
    \label{fig:additionalqualitativeresults}
\end{figure*}

In this supplementary, we first provide background on NeRF and 3DGS in~\cref{sec:background_supp} and dataset
details in~\cref{sec:datasets_supp}.In~\cref{sec:vhullabs} we show the impact of our constraint on
reconstructing the visual hull and study the performance of NeRF models for different reconstructions of the
visual hull. We provide a thorough quantitative comparison against a visual hull-based baseline, VaxNeRF,
in~\cref{sec:vaxnerf_comp}. We briefly note the discrepancies between reported GO performance in
their paper and our work (\cref{sec:go_psnr}). We complement main image quality metrics with LPIPS in~\cref{sec:lpips_supp}. We further study the effect of mask dilation in~\cref{sec:maskdilation}. Finally, we present additional qualitative
results in~\cref{fig:additionalqualitativeresults}.


\section{Background on NeRF and 3DGS}
\label{sec:background_supp}
\subsection{Neural Radiance Fields}
\label{subsec:nerf}
Neural Radiance Field (NeRF) is a 5D continuous representation of a 3D scene that relies on the principles of
classical volumetric rendering. Specifically, given a camera ray $\mathbf{r}(t) = \mathbf{o} + t\mathbf{d}$
and a set of $N$ samples along the ray, an MLP estimates densities $\sigma$ and colors $\mathbf{c}$ at each
sample. The ray's final color $\hat{C}(\mathbf{r})$ is rendered by a finite approximation of the rendering
equation within the near and far bounds $t_n$ and $t_f$ and transmittance $T$:
\begin{equation}
\label{eq:nerf_render}
\hat{C}(\mathbf{r}) = \int_{t_n}^{t_f} T(t) \sigma(t) \, \mathbf{c}(t) \, e^{-\int_{t_1}^{t} \sigma(\tau) \, d\tau} \, dt.
\end{equation}
Some methods starting from IDR~\cite{yariv2020idr} have enforced a silhouette loss to constrain the multi-view
reconstruction for objects as $\mathcal{L}_\text{mask} = -W \log(\hat{W}) - (1-W) \log(1-\hat{W})$, where
$\hat{W}$ is the rendered mask and $W$ is the given ground truth mask.
\subsection{Gaussian Splatting}
\label{subsec:gaussplat}
3D Gaussian Splatting (3DGS)~\cite{kerbl3Dgaussians} is a real-time neural rendering technique that represents
a 3D scene using Gaussian primitives comprising properties such as position $\mathbf{\mu}$, opacity $\alpha$,
covariance $\mathbf{\Sigma}$, and color $\mathbf{c}$. Unlike NeRFs~\cite{mildenhall2020nerf}, which rely on
volumetric sampling and expensive ray marching, 3DGS directly projects and renders Gaussians in a
differentiable way using alpha compositing (given a pixel $\mathbf{p}$ in \cref{eq:gs_render}), enabling fast
and high-quality reconstruction. In this work, we deal with the final rendered image $\hat{C}$ and the
rendered alpha channel $w_i$,
\begin{align}
\label{eq:gs_render}
\hat{C}(\mathbf{p}) &= \sum_{i} w_i \mathbf{c}_i\text{, where  }  w_i = \alpha_i \prod_{j=1}^{i-1} (1 - \alpha_j) \,.
\end{align}

\section{Datasets}
\label{sec:datasets_supp}
\noindent \textbf{ActorsHQ}~\cite{isik2023humanrf} consists of $8$ subjects captured by a $160$-camera fully calibrated rig. We hold out $30$ cameras for evaluation and uniformly sample $5$, $8$, and $12$ training views per subject.
Ground-truth foreground masks are provided with the dataset and used directly for visual hull reconstruction.

\noindent \textbf{MipNeRF360}~\cite{barron2022mipnerf360} We evaluate on the \textit{kitchen},
\textit{bonsai}, and \textit{garden} sequences following the split of
GaussianObject~\cite{yang2024gaussianobject}, with $4$, $6$, and $9$ training images per scene. We use masks and camera poses released by~\cite{yang2024gaussianobject}.

\noindent \textbf{Omni3D}~\cite{brazil2023omni3d} We follow the train/test splits of
GaussianObject~\cite{yang2024gaussianobject} with $4$, $6$, and $9$ training images. We evaluate on $17$
sequences: \textit{backpack\_016, box\_043, broccoli\_003, corn\_007, dinosaur\_006, flower\_pot\_007,
gloves\_009, guitar\_002, hamburger\_012, picnic\_basket\_009, pineapple\_013, sandwich\_003,
suitcase\_006}, \textit{timer\_010, toy\_plane\_005, toy\_truck\_037,} and \textit{vase\_012}. Masks and poses are from~\cite{yang2024gaussianobject}.
\section{VisDom on Visual Hull Reconstruction} \label{sec:vhullabs}
We improve on the visual hull reconstruction accuracy as described in Sec. 4.2, main. To evaluate the effect
of our visual domain constraint, we compare our visual hull reconstruction with (i) the traditional visual hull
algorithm (see Sec. 3.3, main), and (ii) visual hull reconstruction post-processed by removing samples outside
the unit sphere~\cite{kondo2021vaxnerf,yang2024gaussianobject}. We show quantitative results
in~\cref{tab:vhullaccuracy} comparing the visual hull quality against meshes from the ActorsHQ dataset using
the symmetric L2 Chamfer distance:
\begin{equation*}
    \sum_{x_i \in \, \text{T}} \min_{y_i \in \, \text{R}} \|x_i-y_i \|_2^2 + \sum_{y_j \in \, \text{R}}  \min_{x_j \in \, \text{T}} \|x_j-y_j \|_2^2 \,,
\end{equation*}
where $x_i$ and $x_j$ are points on the target shape (T), and $y_i$ and $y_j$ are points on the reconstructed
shape (R). As demonstrated by the superior numbers, our proposed geometric constraint enables high-quality
visual hull reconstruction from as few as $4$ views. Moreover, the L2 ground truth to reconstruction accuracy
on an average is $0.35$, $0.40$, and $1.31$, respectively, with ours, with unit sphere and traditional.
These results further corroborate the merits of our constraint compared with the unit sphere heuristic. 
In~\cref{fig:vhull_abs}, we show the qualitative results of the visual hull reconstruction for different
numbers of cameras.
\begin{table}[ht]
    \centering
    \begin{tabular}{c|c|c|c}
    & \multicolumn{3}{c}{Visual hull} \\
     & with ours & with unit sphere & traditional \\ \cline{2-4}
    Cameras & \multicolumn{3}{c}{$10^3 \times$ L2 Symm. chamfer distance ($\downarrow$)} \\ \hline
$5$ & $\mathbf{1.05}$ & $683.25$ & $4366.50$ \\
$8$ & $\mathbf{1.34}$ & $523.41$ & $4847.00$ \\
$12$ & $\mathbf{0.64}$ & $266.62$ & $5293.00$ \\ \hline
Mean & $\mathbf{1.10}$ & $537.59$ & $4578.44$ \\
    \end{tabular}
    \caption{Reconstruction accuracy of the visual hull obtained from different camera configurations from the
    ActorsHQ dataset. The best results per row are shown in \textbf{bold}.}
    \label{tab:vhullaccuracy}
\end{table}
\begin{figure}[ht!]
    \centering
    \scalebox{0.8}{
    \includegraphics[width=0.72\linewidth]{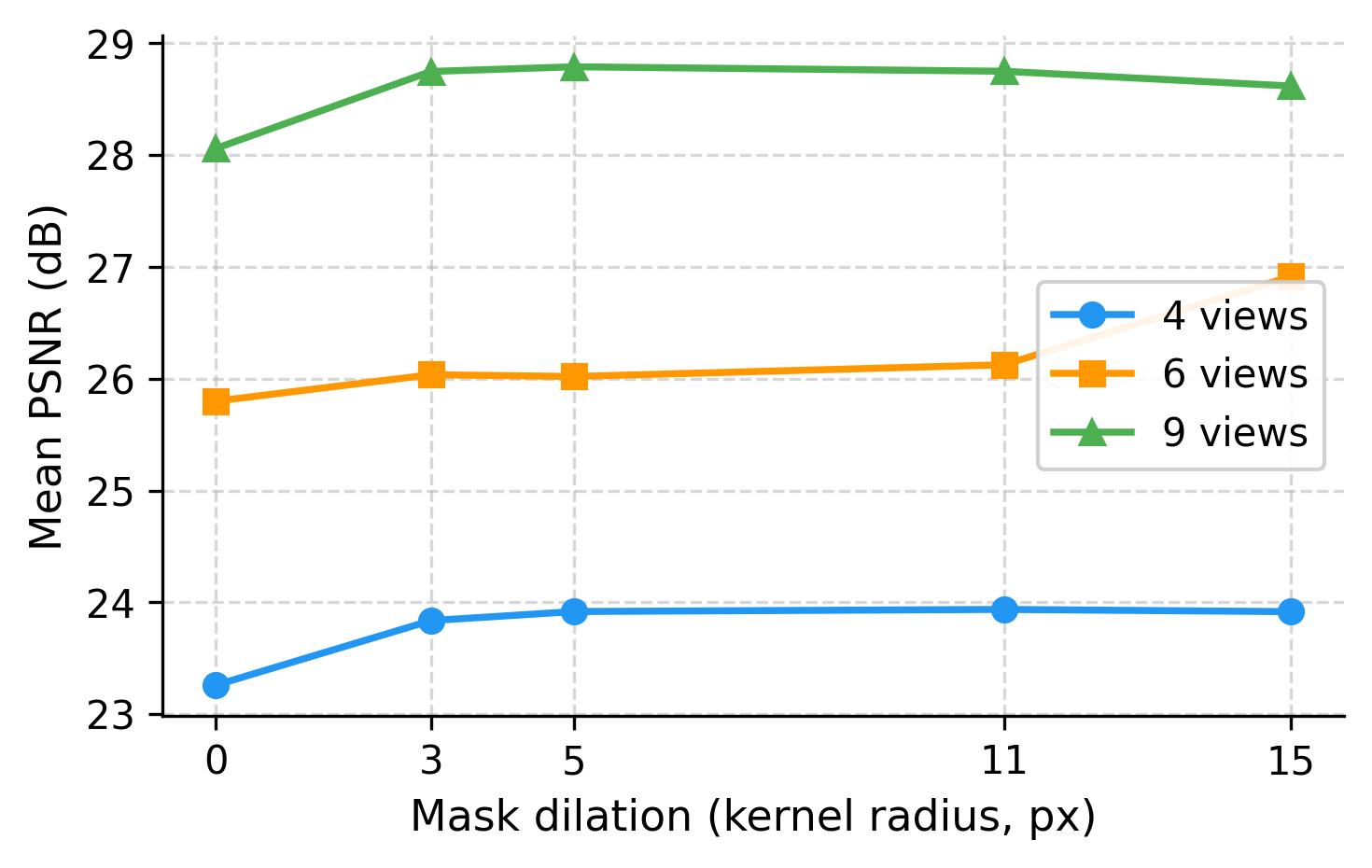}}
    \caption{Effect of mask dilation on ZipNeRF~+~VisDom for MipNeRF360 dataset, measured in mean PSNR (dB) averaged over \textit{bonsai}, \textit{garden}, and \textit{kitchen}. The $r{=}0$ point corresponds to the results reported in the main paper. Dilation improves the performance over the no-dilation baseline, suggesting the benefit of applying small padding to the mask boundary as mitigation for their multi-view inconsistencies.\vspace{-3mm}}
    \label{fig:maskdilationsens}
\end{figure}
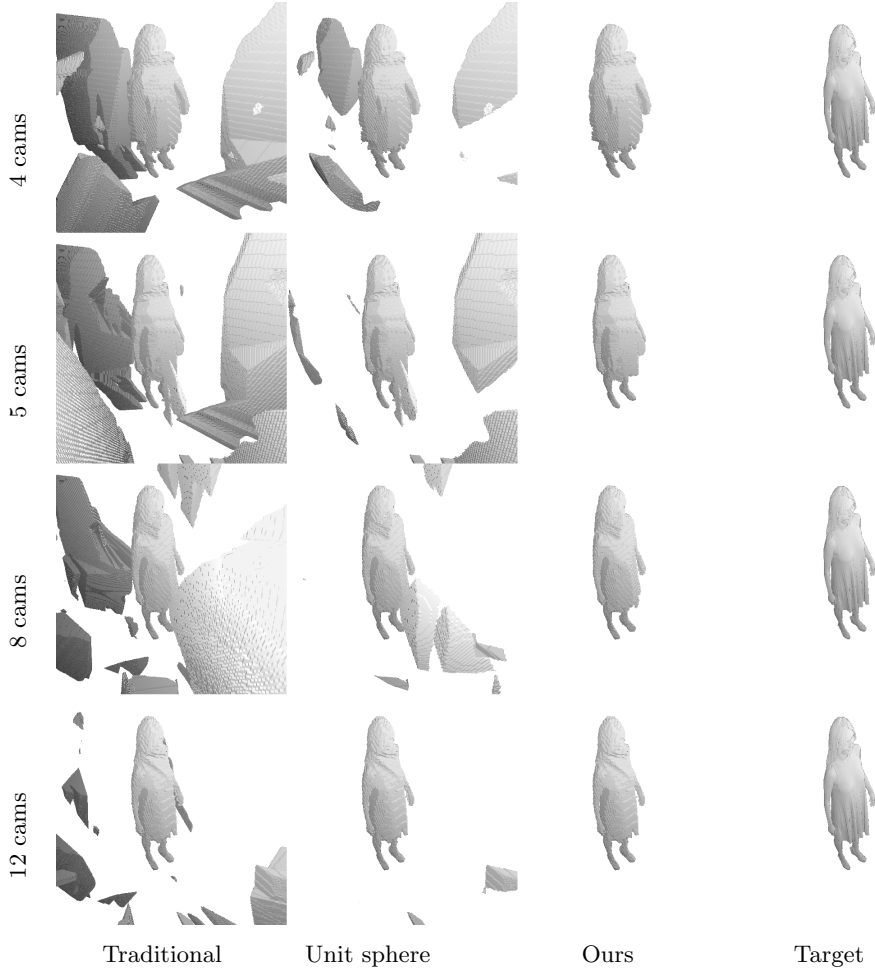
\begin{figure}[ht]
    \centering
    \begin{overpic}[width=1.0\linewidth,tics=5,]
        {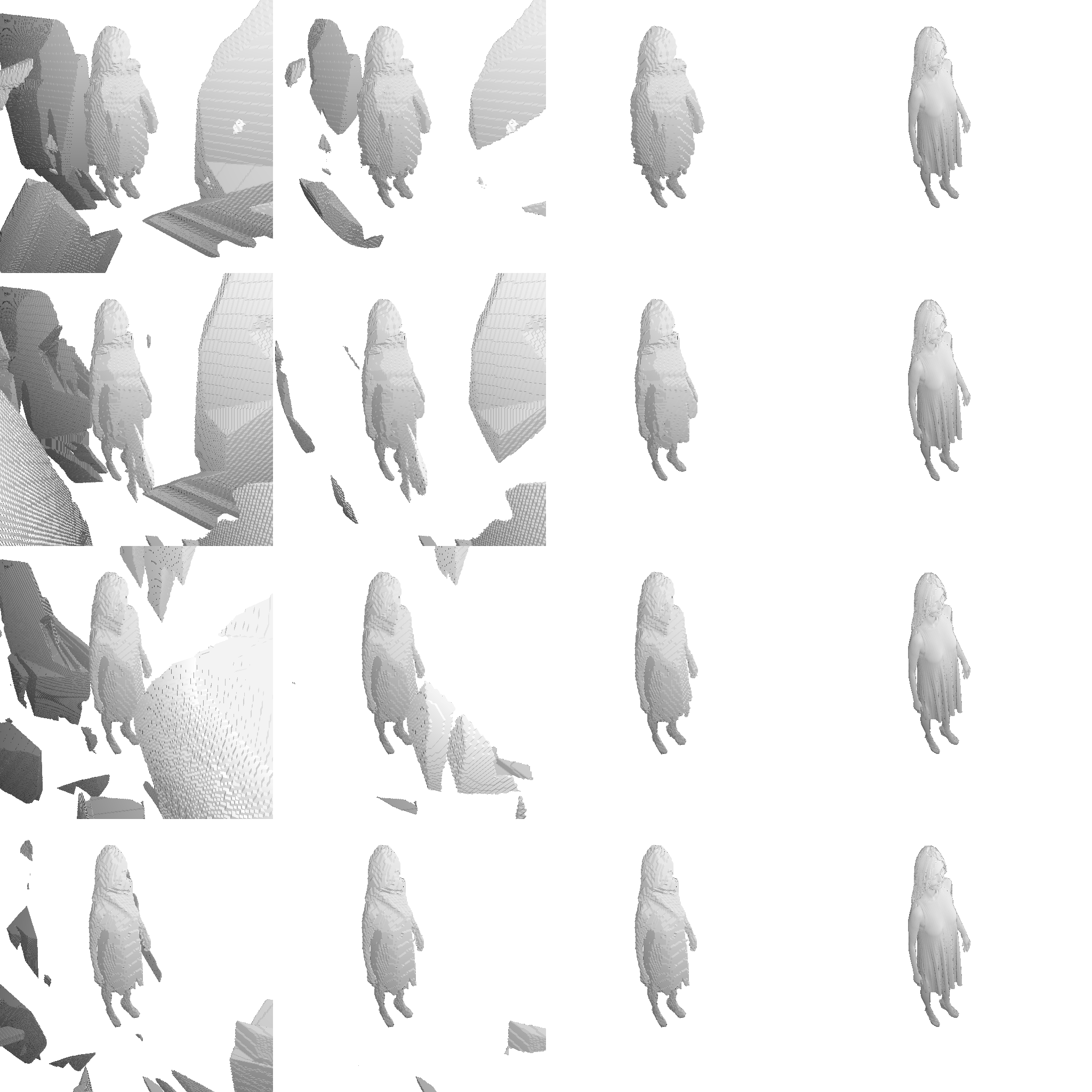}
        
        \put(-5,80){\rotatebox{90}{{\small $4$ cams}}}
        \put(-5,55){\rotatebox{90}{{\small $5$ cams}}}
        \put(-5,30){\rotatebox{90}{{\small $8$ cams}}}
        \put(-5,5){\rotatebox{90}{{\small $12$ cams}}}
        
        \put( 5,-4){{\small Traditional}}
        \put(27,-4){{\small Unit sphere}}
        \put(57,-4){{\small Ours}}
        \put(80,-4){{\small Target}}
    \end{overpic}
    \vspace{0.2mm}
    \caption{Qualitative results of the visual hull reconstructed using different algorithms, from left to
    right: traditional, unit sphere heuristic, and with our VisDom constraint.}
    \label{fig:vhull_abs}
\end{figure}

When there are many artifacts around the reconstructed shape (see left column of ~\cref{fig:vhull_abs}), the
right part of the equation would be too large, which we observe in the symmetric distance
in~\cref{tab:vhullaccuracy}. In~\cref{tab:vhullablations}, we show the error measured from target to
reconstruction to evaluate the shape quality apart from the unwanted regions. As corroborated numerically, our
visual domain constraint helps carve additional regions compared to the traditional visual hull algorithm and unit
sphere heuristic. Thus, reconstructing a visual hull with our constraint leads to better results, in general,
not just in terms of removing unwanted regions away from the shape.

\begin{figure}
    \centering        
    \begin{overpic}[width=1.0\linewidth,tics=5,]
        {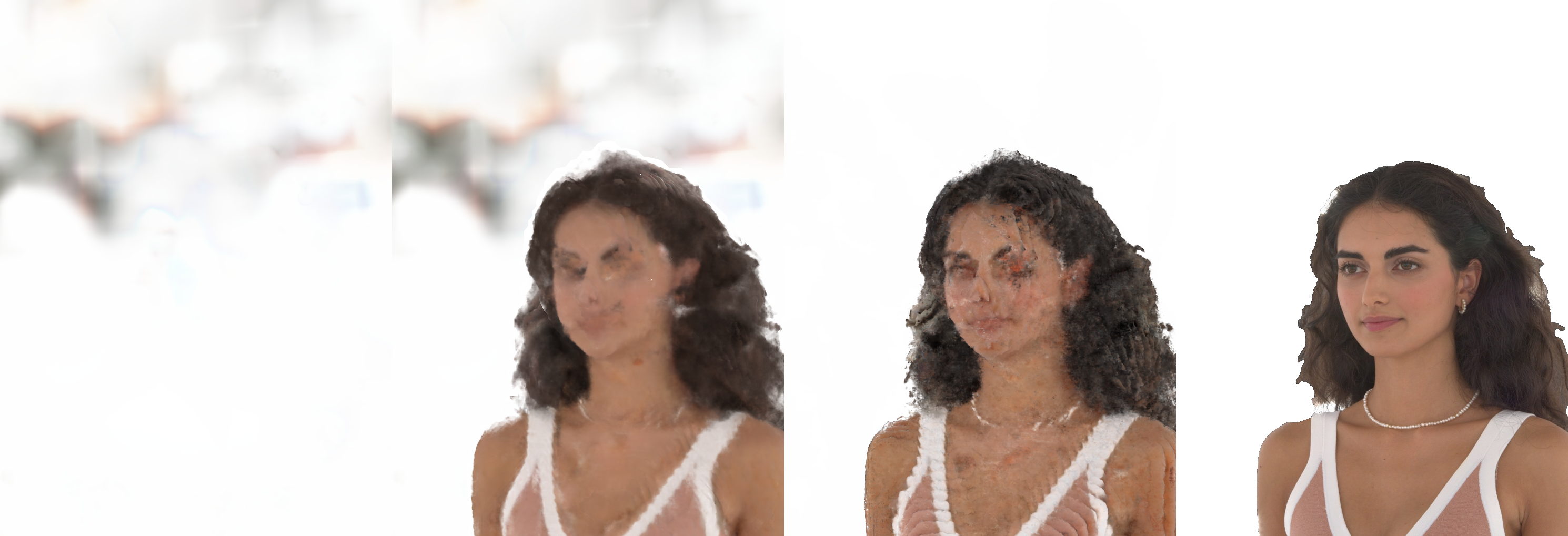}
    \put(-4,10){\rotatebox{90}{\small ActorsHQ}}
    \end{overpic}
    
    \begin{overpic}[width=1.0\linewidth,tics=5,]
        {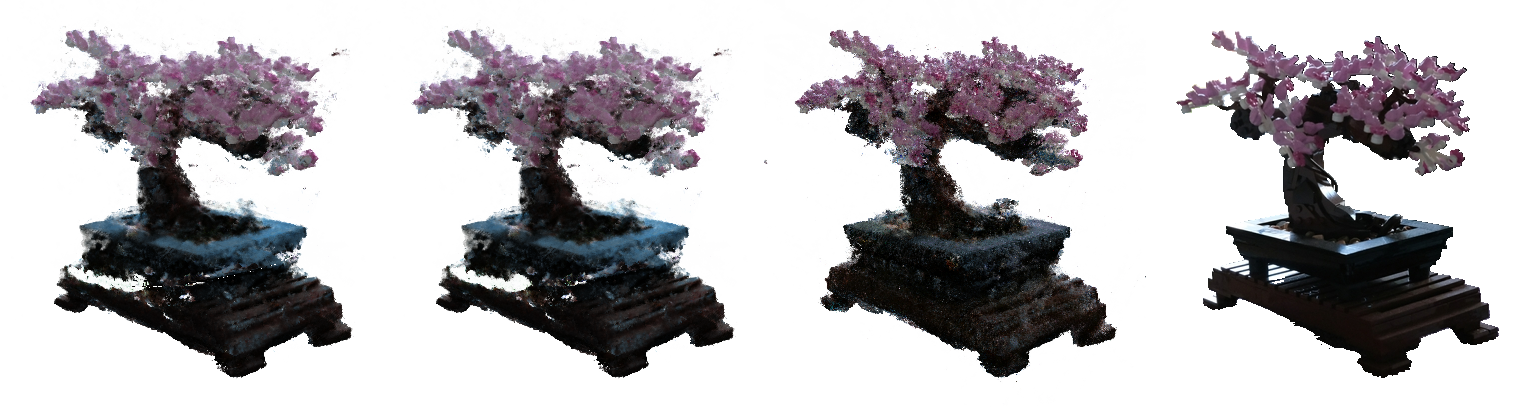}
    \put(-4,3){\rotatebox{90}{\small MipNeRF360}}
    \end{overpic}

    \begin{overpic}[width=1.0\linewidth,tics=5,]{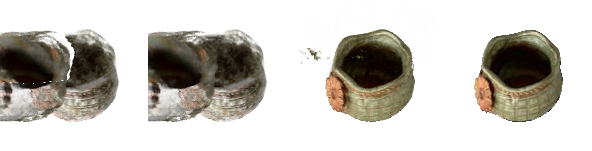}
    
    \put(-4,5){\rotatebox{90}{\small Omni3D}}
    \put(3,-2){\small VaxNeRF}
    \put(28,-2){\small VaxNeRF$^*$}
    \put(55,-2){\small Instant-NGP}
    \put(58,-5){\small w ours}
    \put(78,-2){\small Ground truth}
    \end{overpic}
    \vspace{0.5mm}
    \caption{Qualitative comparisons with VaxNeRF and VaxNeRF$^*$ (visual hull used for inference) models on
    $12$, $9$, and $9$ cameras respectively for the three datasets.}
    \label{fig:vaxnerfqual}
\end{figure}

\begin{figure*}
    \centering

    \begin{overpic}[width=0.8\linewidth,tics=5,]
        {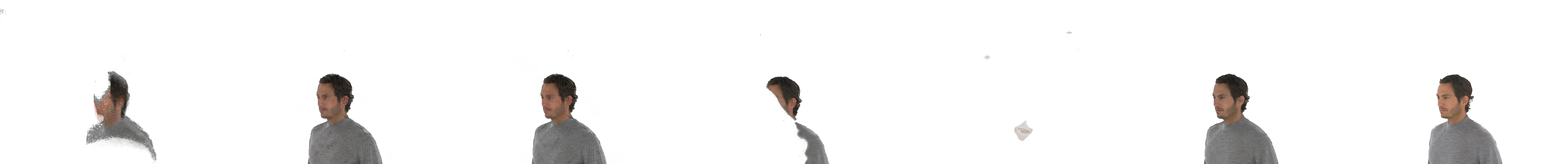}
       \put(-2,-1){\rotatebox{90}{\tiny $5$ cams}}
    \end{overpic}
    
    \begin{overpic}[width=0.8\linewidth,tics=5,]
        {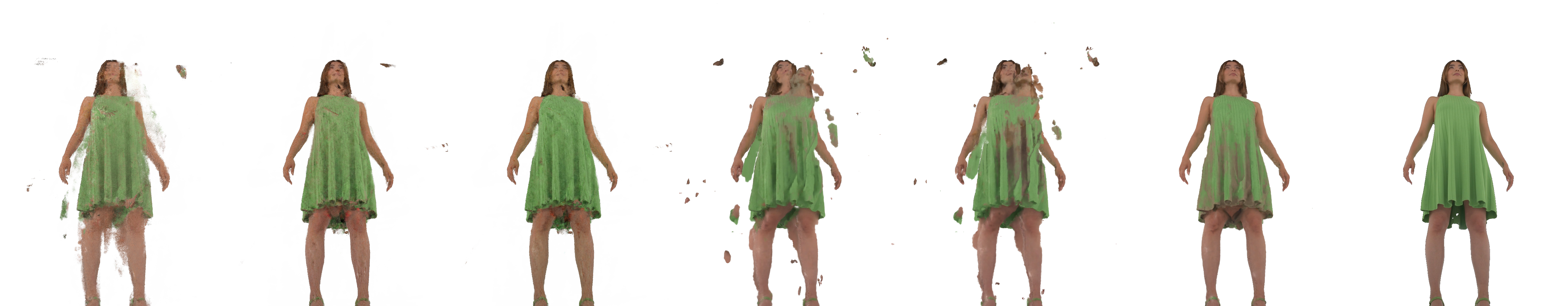}
       \put(-2,3){\rotatebox{90}{\tiny $8$ cams}}
       \put(-6,0){\rotatebox{90}{\small ActorsHQ}}
    \end{overpic}
    \begin{overpic}[width=0.8\linewidth,tics=5,]
        {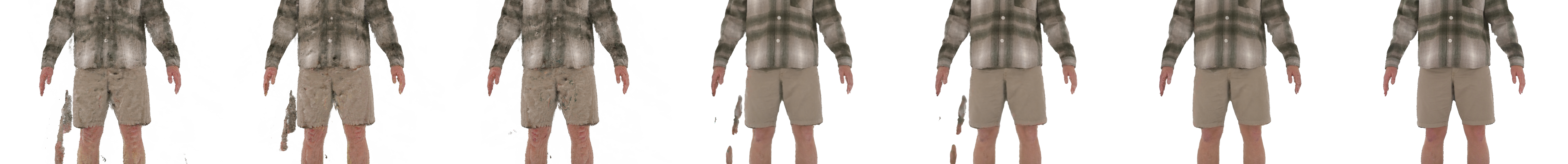}
       \put(-2,1){\rotatebox{90}{\tiny $12$ cams}}
       \put(0,-1){\linethickness{0.25mm}\color{black}\line(1,0){100}}%
    \end{overpic}
    \vspace{2mm}
    
    \begin{overpic}[width=0.8\linewidth,tics=5,]
        {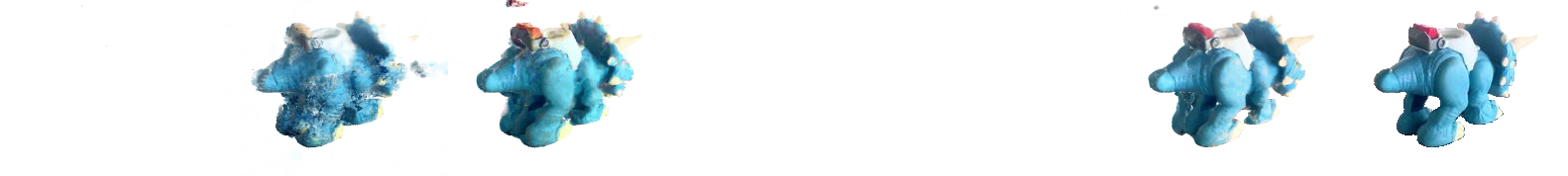}
       \put(-2,2){\rotatebox{90}{\tiny $4$ cams}}
    \end{overpic}
    \begin{overpic}[width=0.8\linewidth,tics=5,]
        {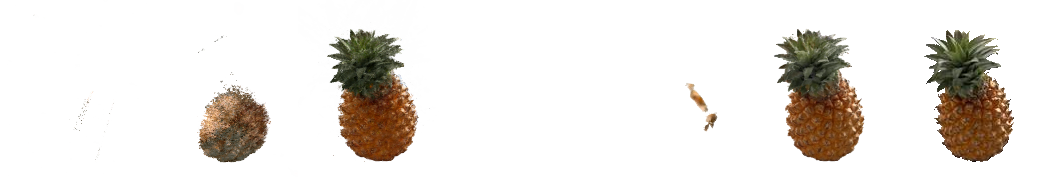}
       \put(-6,0){\rotatebox{90}{\small Omni3D}}
       \put(-2,2){\rotatebox{90}{\tiny $6$ cams}}
    \end{overpic}
    \begin{overpic}[width=0.8\linewidth,tics=5,]
        {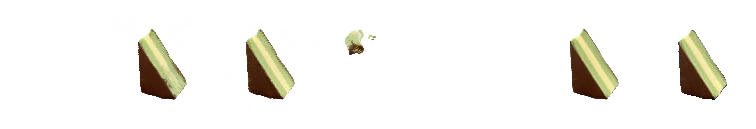}
       \put(-2,4){\rotatebox{90}{\tiny $9$ cams}}
       \put(0,1){\linethickness{0.25mm}\color{black}\line(1,0){100}}%
    \end{overpic}

    \begin{overpic}[width=0.8\linewidth,tics=5,]
        {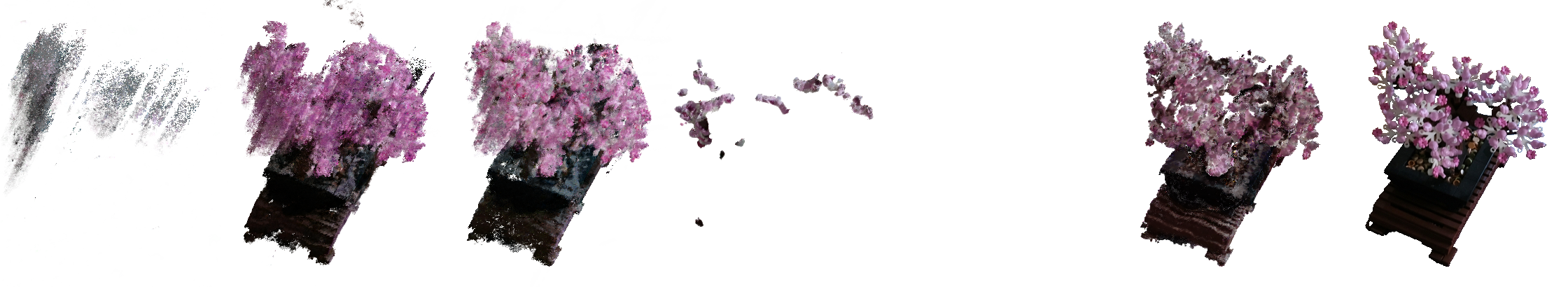}
       \put(-2,5){\rotatebox{90}{\tiny $4$ cams}}
    \end{overpic}
    \begin{overpic}[width=0.8\linewidth,tics=5,]
        {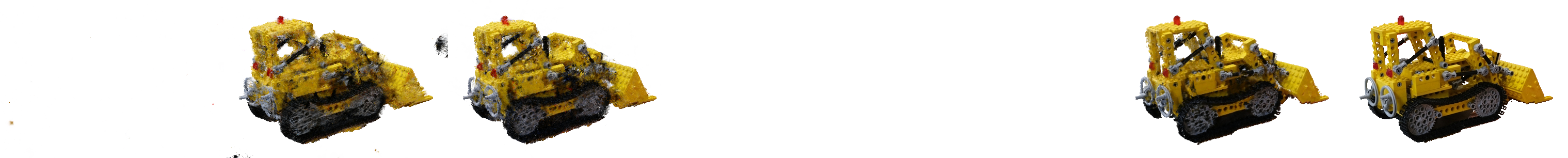}
       \put(-2,2){\rotatebox{90}{\tiny $6$ cams}}
       \put(-6,-2){\rotatebox{90}{\small Mip360}}
    \end{overpic}
    \begin{overpic}[width=0.8\linewidth,tics=5,]
        {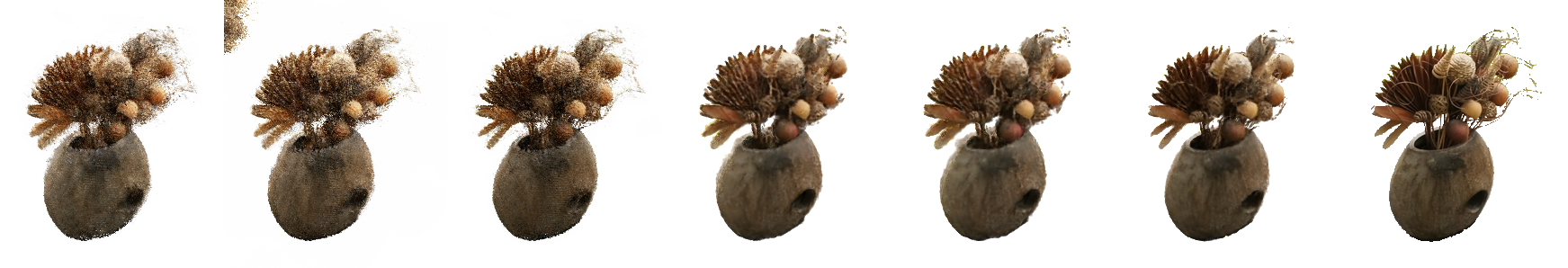}
       \put(-2,4){\rotatebox{90}{\tiny $9$ cams}}
        
       \put(0,-1){\tiny traditional}
       \put(13,-1){\tiny unit sphere}
       \put(15,-3.5){\small Instant-NGP}
       \put(31,-1){\tiny VisDom}

       \put(42.5,0){\linethickness{0.25mm}\color{black}\line(0,1){130}}%
       
       \put(44,-1){\tiny traditional}
       \put(56.5,-1){\tiny unit sphere}
       \put(57,-3.5){\small ZipNeRF}
       \put(75,-1){\tiny VisDom}
       
       \put(85.5,0){\linethickness{0.25mm}\color{black}\line(0,1){130}}%
       
       \put(86.5,-1){\small Ground}
       \put(88,-3.5){\small truth}
    \end{overpic}
    \vspace{5mm}
    \caption{Novel view synthesis from ZipNeRF and Instant-NGP models when trained with our constraint based
    on different visual hull algorithms. When trained with the visual hull reconstructed with our constraint,
    the models perform the best in most settings.}
    \label{fig:vhullAblationsNerfs}
\end{figure*}

Although the constraint helps reconstruct more accurate shapes based on silhouettes, it is still unclear if
the better shape means a better reconstruction with our proposed geometric constraint on NeRFs. Towards this,
we evaluate the performance of our constraint on ZipNeRF and Instant-NGP with different visual hulls
reconstructed using the traditional visual hull algorithm, culling outside the unit sphere heuristic (unit
sphere), and with ours. As can be seen from ~\cref{tab:vhullAblationsNerfs} and
~\cref{fig:vhullAblationsNerfs}, our constraint performs best across all settings. The traditional hull
leaves large spurious regions that severely degrade NeRF training, and while the unit-sphere heuristic helps
when the object is well centered, it still underperforms our data-driven constraint. Unlike the unit sphere,
VisDom does not enforce central object placement and can effectively remove spurious regions outside the
object.

\begin{table}[ht]
    \centering
    \begin{tabular}{c|c|c|c}
    & \multicolumn{3}{c}{Visual hull} \\
     & with ours & with unit sphere & traditional \\ \cline{2-4}
    Cameras & \multicolumn{3}{c}{$10^3 \times$ L2 target $\rightarrow$ rec. chamfer distance ($\downarrow$)} \\ \hline
$5$  &  $\textbf{0.38}$ & $0.44$ & $1.34$ \\
$8$  &  $\textbf{0.37}$ & $0.40$ & $1.57$ \\
$12$ &  $\textbf{0.21}$ & $\textbf{0.21}$ & $1.09$ \\ \hline
Mean &  $\textbf{0.35}$ & $0.40$ & $1.31$ \\ 
    \end{tabular}
    \caption{Reconstruction accuracy, target shape to the reconstructed shape, of the visual hull obtained
    from different camera configurations from the ActorsHQ dataset. The best results per row are shown in
    \textbf{bold}.}
    \label{tab:vhullablations}
\end{table}

\begin{table}[ht]
    \centering
    \setlength\tabcolsep{1.5pt}
\scalebox{0.9}{
    \begin{tabular}{c|c|c|c|c||c|c|c}
& \multirow{2}{*}{Cameras} & \multicolumn{3}{c||}{Instant-NGP} & \multicolumn{3}{c}{ZipNeRF} \\  \cline{3-8}
 & & traditional & unit sphere & +VD & traditional & unit sphere & +VD \\ \hline
 & & \multicolumn{6}{c}{PSNR ($\uparrow$)} \\ \cline{3-8}
\multirow{4}{*}{\rotatebox{90}{\small ActorsHQ}} & $5$ & $5.50$ & $21.00$ & $\mathbf{23.53}$ &  $6.30$ & $6.19$ & $\mathbf{24.55}$ \\
& $8$ & $20.72$ & $21.61$ & $\mathbf{23.35}$ & $19.97$ & $19.99$ & $\mathbf{26.72}$ \\
& $12$ & $20.70$ & $21.80$ & $\mathbf{25.67}$ & $23.48$ & $23.44$ & $\mathbf{28.61}$ \\ \cline{2-8}
& Mean & $15.64$ & $21.47$ & $\mathbf{24.18}$ & $16.58$ & $16.54$ & $\mathbf{26.63}$ \\ \hline
\multirow{4}{*}{\rotatebox{90}{Omni3D}}& $4$ & $16.83$ & $18.50$ & $\mathbf{27.44}$ & $16.87$ & $16.87$ & $\mathbf{29.49}$ \\
& $6$ & $16.77$ & $19.71$ & $\mathbf{29.67}$ & $16.91$ & $16.90$ & $\mathbf{32.28}$ \\
& $9$ & $16.60$ & $20.87$ & $\mathbf{31.06}$ & $16.96$ & $16.90$ & $\mathbf{35.21}$ \\ \cline{2-8}
& Mean & $16.73$ & $19.69$ & $\mathbf{29.39}$ & $16.91$ & $16.89$ & $\mathbf{32.33}$ \\  \hline
\multirow{4}{*}{\rotatebox{90}{\shortstack{Mip360}}} & $4$ & $17.05$ & $19.59$ & $\mathbf{22.15}$ & $17.28$ & $17.18$ & $\mathbf{24.10}$ \\
& $6$ & $17.58$ & $21.43$ & $\mathbf{24.14}$ & $17.58$ & $18.03$ & $\mathbf{25.80}$ \\
& $9$ & $21.22$ & $22.00$ & $\mathbf{25.43}$ & $19.87$ & $20.06$ & $\mathbf{28.06}$ \\ \cline{2-8}
& Mean & $18.61$ & $21.01$ & $\mathbf{23.91}$ & $18.24$ & $18.42$ & $\mathbf{25.99}$ \\

    \end{tabular}
}
    \caption{Novel view synthesis accuracy of ZipNeRF and Instant-NGP models when trained with our VisDom
    constraint based on different visual hull algorithms. The best performance for each method, dataset and
    camera configuration is shown in \textbf{bold}. As it can be observed from the evaluations, the model
    performance indeed depends on the accuracy of the visual hull. Therefore, when trained with the visual
    hull reconstructed with our constraint, the models perform the best in most settings, boosting traditional
    visual hull performance in some challenging scenarios (e.g. ActorsHQ, 5 cameras) by over \textbf{200\%}.}
    \label{tab:vhullAblationsNerfs}
\end{table}

\begin{table}[ht]
    \centering
    \setlength\tabcolsep{1.5pt} 
    \begin{tabular}{c|c|c|c|c}
    \multirow{2}{*}{Dataset} & \multirow{2}{*}{Cameras} & VaxNeRF & VaxNeRF$^*$ & INGP + VD \\ \cline{3-5}
    & & \multicolumn{3}{c}{PSNR $(\uparrow)$} \\ \cline{2-5} 
    \multirow{4}{*}{\rotatebox{90}{\small ActorsHQ}} &$5$ & $19.25$ & $22.90$ & $\textbf{23.53}$ \\
    &$8$ & $21.35$ & $\textbf{24.38}$ & $23.35$ \\
    &$12$ & $23.04$ & $\textbf{26.27}$ & $25.67$ \\ \cline{2-5}
    &Mean & $21.21$ & $\textbf{24.52}$ & $24.18$ \\ \hline
    \multirow{4}{*}{\rotatebox{90}{Omni3D}}& $4$ & $18.33$ & $18.35$ & $\textbf{27.44}$ \\
    & $6$ & $19.52$ & $19.60$ & $\textbf{29.67}$ \\
    & $9$ & $20.76$ & $20.91$ & $\textbf{31.06}$ \\  \cline{2-5}
    & Mean & $19.54$ & $19.62$ & $\textbf{29.39}$ \\ \hline
    \multirow{4}{*}{\rotatebox{90}{\shortstack{MipNeRF\\ 360}}}& $4$ & $19.19$ & $19.45$ & $\textbf{22.15}$ \\
    & $6$ & $21.28$ & $21.68$ & $\textbf{24.14}$ \\
    & $9$ & $22.32$ & $22.57$ & $\textbf{25.43}$ \\  \cline{2-5}
    & Mean & $20.93$ & $21.23$ & $\textbf{23.91}$ \\
    \end{tabular}
    \caption{Quantitative comparisons with VaxNeRF and VaxNeRF$^*$ (visual hull used for inference) models on
    the three datasets from the main paper. The best results for each row are shown in \textbf{bold}. For the
    ActorsHQ dataset, as the quality of the visual hull is better with more cameras, the VaxNeRF$^*$ model
    performs better while using the visual hull during the inference. Instant-NGP with our VisDom constraint (INGP + VD), on the
    other hand, is competitive. However, on other datasets, Instant-NGP with our VisDom constraint outperforms
    even the VaxNeRF$^*$ due to our improved quality of the visual hull and our geometric constraint.}
    \vspace{-3mm}
    \label{tab:vaxnerfquant}
\end{table}

\section{Additional Comparisons Against VaxNeRF, a Visual Hull-based Approach}
\label{sec:vaxnerf_comp}
In the main paper, for a fair comparison, we trained VaxNeRF with a recent model, Instant-NGP, instead of the
vanilla NeRF. We compared the results by rendering the VaxNeRF model without additional information during
inference. However, in VaxNeRF, the idea is to learn the representation inside the visual hull for
acceleration. Therefore, the model is only rendered inside the visual hull, even during the inference. To make
a fairer comparison with the VaxNeRF model, we show quantitative (see ~\cref{tab:vaxnerfquant}) and
qualitative (see ~\cref{fig:vaxnerfqual}) comparisons in both settings, without visual hull during inference
(VaxNeRF), and with visual hull during inference (VaxNeRF$^*$). For ease, we show the numbers of VaxNeRF and
Instant-NGP with ours from the main paper.

As expected, the VaxNeRF model, when rendered without a visual hull, results in artifacts as the model ignores
the regions during training.  While the renderings of VaxNeRF$^*$ are slightly sharper than the VaxNeRF model,
the method still suffers from artifacts due to its visual hull heuristic (regions culled outside the unit
sphere) that is not always suitable as discussed in ~\cref{sec:vhullabs}.
\section{Note on GaussianObject Reported PSNR Values}
\label{sec:go_psnr}
We discovered a dissimilarity in GO’s PSNR calculation
\href{https://github.com/chensjtu/GaussianObject/blob/91048a5dd3f9313e5bca8ecca651bdbac5099836/render.py#L168}{(render.py
L166-168)} with the commonly adopted global-PSNR evaluation~\cite{kerbl3Dgaussians}. In GO, image quality is
estimated by computing MSE for each channel separately and then averaging PSNR value across three color
channels (average-per-channel). The estimation is biased as it can overweigh low-error channels. In our
evaluation protocol we adopt a global standard, where all pixels and channels are treated equally. Using GO’s
original evaluation code, we are able to reproduce their reported metric values. Nonetheless, to stay
consistent with the unbiased practices we report GO's performance in Table 2 (main paper) according to the
global PSNR values.

\section{Perceptual Quality (LPIPS)}
\label{sec:lpips_supp}
To complement the PSNR and SSIM results in Table~2 (main), we report LPIPS~\cite{zhang2018perceptual} for the
general-purpose NeRF and 3DGS methods in~\cref{tab:lpips_supp}. The perceptual metric tells the same story:
VisDom yields large gains for the NeRF methods (e.g., Instant-NGP $22.0{\rightarrow}9.5$ and ZipNeRF
$16.2{\rightarrow}7.4$ at 4 views on MipNeRF360), turning previously unusable reconstructions into perceptually
faithful ones. For 3DGS-GO, whose GaussianObject initialization already benefits from depth and silhouette
priors, VisDom remains on par in LPIPS while improving PSNR and SSIM (Table~2, main), consistent with its role
as a complementary geometric regularizer rather than a competing appearance prior.

\begin{table*}[ht]
\centering
\setlength\tabcolsep{2pt}
\scalebox{0.92}{
\begin{tabular}{l|c|ccc|ccc|cc}
\multirow{2}{*}{Dataset} & \multirow{2}{*}{Views}
& \multicolumn{3}{c|}{INGP~\cite{mueller2022instant}}
& \multicolumn{3}{c|}{ZipNeRF~\cite{barron2023zipnerf}}
& \multicolumn{2}{c}{3DGS-GO~\cite{yang2024gaussianobject}} \\
\cmidrule(lr){3-5} \cmidrule(lr){6-8} \cmidrule(lr){9-10}
& & van. & +mask & +VD & van. & +mask & +VD & van.* & +VD \\
\midrule
& & \multicolumn{8}{c}{LPIPS $\times 10^2$ $\downarrow$} \\
\cmidrule(lr){3-10}
\multirow{3}{*}{\rotatebox[origin=c]{90}{\small Mip360}}
 & 4 & 22.021 & 23.880 & \textbf{9.452} & 16.225 & 18.749 & \textbf{7.448} & \textbf{5.890} & 5.942 \\
 & 6 & 21.361 & 17.831 & \textbf{8.354} & 19.462 & 19.633 & \textbf{4.377} & \textbf{4.269} & 4.569 \\
 & 9 & 20.388 & 13.720 & \textbf{8.714} & 20.021 & 13.696 & \textbf{3.148} & \textbf{3.378} & 3.814 \\
\midrule
\multirow{3}{*}{\rotatebox[origin=c]{90}{\small Omni3D}}
 & 4 & 14.237 & 15.194 & \textbf{5.267} & 12.977 & 14.361 & \textbf{3.115} & 2.683 & \textbf{2.627} \\
 & 6 & 12.737 & 11.142 & \textbf{4.503} & 12.611 & 8.759 & \textbf{1.863} & \textbf{1.899} & 1.907 \\
 & 9 & 13.987 & 10.112 & \textbf{4.358} & 8.130 & 5.208 & \textbf{1.280} & \textbf{1.470} & 1.514 \\
\midrule
\multirow{3}{*}{\rotatebox[origin=c]{90}{\small ActHQ}}
 & 5  & 33.910 & 31.953 & \textbf{14.343} & 24.812 & 28.535 & \textbf{12.905} & \textbf{11.625} & 11.713 \\
 & 8  & 34.655 & 28.736 & \textbf{14.782} & 23.381 & 28.119 & \textbf{9.517} & \textbf{10.209} & 10.506 \\
 & 12 & 20.302 & 13.615 & \textbf{13.282} & 25.613 & 9.028 & \textbf{8.980} & \textbf{9.371} & 9.753 \\
\end{tabular}
}
\caption{Perceptual quality (LPIPS $\times 10^2$, lower is better) of VisDom on general-purpose NeRF and 3DGS
methods, complementing the PSNR/SSIM results in Table~2 (main). Best variant per method group is in
\textbf{bold}. \emph{van.*} - the initialization stage of GaussianObject~\cite{yang2024gaussianobject}. VisDom
yields large perceptual gains for the NeRF methods, while remaining on par with the already depth- and
silhouette-regularized 3DGS-GO initialization.}
\label{tab:lpips_supp}
\end{table*}

\section{Effect of Mask Dilation}
\label{sec:maskdilation}

Our visual hull reconstruction relies on binary foreground silhouettes. In practice, mask boundaries may be slightly inaccurate, and small morphological perturbations can affect the tightness of the reconstructed hull. We study the impact of mask dilation on ZipNeRF with our VisDom constraint applied during visual hull reconstruction. Specifically, we dilate the foreground masks with a square structuring element of varying kernel radius $r \in \{0, 3, 5, 11, 15\}$ pixels on the MipNeRF360 dataset (\textit{bonsai}, \textit{garden}, \textit{kitchen}) for all three training-view configurations.

\cref{fig:maskdilationsens} plot demonstrates the mean PSNR averaged across scenes for each view count, with $r{=}0$ corresponding to the ZipNeRF+VD results reported in the main paper. Dilation consistently improves performance across all view counts. For $4$ views, performance rises from $r{=}0$ to $r{=}3$ ($+0.58\,$dB) and then plateaus, suggesting that a small safety margin around the mask boundary is sufficient to prevent the visual hull from clipping the true object when only four silhouettes constrain the reconstruction. For $9$ views, a similar early plateau is observed with a moderate gain of up to $+0.73\,$dB at $r{=}5$. The $6$-view setting shows the largest absolute gain ($+1.11\,$dB at $r{=}15$), with performance continuing to increase as dilation kernel radius increases. 
We use no dilation ($r = 0$) as reported in the main paper for simplicity and to avoid dataset-specific tuning; moderate dilation (e.g $r{=}3$) can be applied when further performance gains are desired.

\end{document}